\documentclass[runningheads]{llncs}

\usepackage{eccv}

\usepackage{eccvabbrv}

\usepackage{graphicx}
\usepackage{booktabs}
\usepackage{amsmath}
\usepackage{multirow} 
\usepackage{makecell}

\usepackage[accsupp]{axessibility}  % Improves PDF readability for those with disabilities.

\usepackage{hyperref}

\usepackage{orcidlink}

\begin{document}

% ---------------------------------------------------------------
\title{Do Text-free Diffusion Models Learn Discriminative Visual Representations?}

\author{%
Soumik Mukhopadhyay$^{*}$\inst{1}\orcidlink{0009-0003-4134-1805} \and Matthew Gwilliam$^{*}$\inst{1}\orcidlink{0000-0001-9826-6285} \and Yosuke Yamaguchi$^{\dagger}$\inst{2}\orcidlink{0000-0002-0816-2931} \and Vatsal Agarwal$^{\dagger}$\inst{1}\orcidlink{0009-0004-9470-198X} \and
Namitha Padmanabhan\inst{1}\orcidlink{0009-0004-0257-8831} \and Archana Swaminathan\inst{1}\orcidlink{0009-0005-1412-9790} \and
Tianyi Zhou\inst{1}\orcidlink{0000-0001-5348-0632} \and Jun Ohya\inst{2}\orcidlink{0000-0001-7148-4127}  \and Abhinav Shrivastava\inst{1}\orcidlink{0000-0001-8928-8554}
}

\authorrunning{S.~Mukhopadhyay et al.}

\institute{%
  University of Maryland, College Park, MD, USA 
  % \\
  % \email{\{soumik,mgwillia,vatsalag,namithap,archswam,tianyi\}@umd.edu, abhinav@cs.umd.edu}
  \and
  Waseda University, Tokyo, Japan 
  % \\
  % \email{yoyo-015@toki.waseda.jp, ohya@waseda.jp}
}

% ---------------------------------------------------------------
\def\thefootnote{*}\footnotetext{Equal contribution. Corresponding authors \email{\{soumik,mgwillia\}@umd.edu}}
\def\thefootnote{$\dagger$}\footnotetext{Equal contribution.}
\maketitle

\begin{abstract}

Diffusion models have proven to be state-of-the-art methods for generative tasks.  
These models involve training a U-Net to iteratively predict and remove noise, and the resulting model can synthesize high-fidelity, diverse, novel images. 
However, text-free diffusion models have typically not been explored for discriminative tasks. 
In this work, we take a pre-trained unconditional diffusion model and analyze its features post hoc.  
We find that the intermediate feature maps of the pre-trained U-Net are diverse and have hidden discriminative representation properties.
To unleash the potential of these latent properties of diffusion models, we present novel aggregation schemes. 
Firstly, we propose a novel attention mechanism for pooling feature maps and further leverage this mechanism as DifFormer, a transformer feature fusion of different diffusion U-Net blocks and noise steps. 
Next, we also develop DifFeed, a novel feedback mechanism tailored to diffusion. 
We find that diffusion models are better than GANs, and, with our fusion and feedback mechanisms, can compete with state-of-the-art representation learning methods for discriminative tasks -- image classification with full and semi-supervision, transfer for fine-grained classification, object detection, and semantic segmentation.
Our project \href{https://mgwillia.github.io/diffssl/}{website} and \href{https://github.com/soumik-kanad/diffssl}{code} are available publicly. 
  
  
\end{abstract}

\section{Introduction}

\label{sec:intro}

\begin{figure}[h!]
    \centering
    \includegraphics[width=\linewidth]{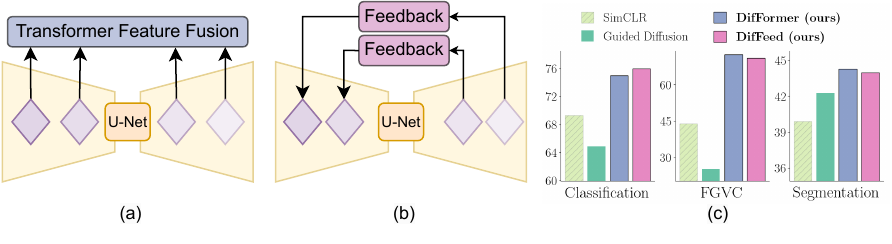}

    \caption{An overview of our method and results. We propose that out-of-the-box pre trained unconditional diffusion models inherently have discriminative properties that automatically make them unified self-supervised image representation learners, with impressive performance not only for generation, but also for discrimination. We improve on the promising results of out-of-the-box diffusion classifiers with our (a) fusion-based DifFormer, and (b) feedback-based DifFeed methods for intelligently utilizing the unique features of diffusion models. (c) We report exciting performances of our methods on multiple downstream benchmarks.}
    \label{fig:teaser_fig}

\end{figure}

For generative tasks, a deep learning model seeks to synthesize or edit parts of images, while for discriminative tasks, it learns to label images or parts of images. 
Traditionally, it has been believed that the representation learned for one is not well-suited for the other because generative models rely on representations that capture low-level (pixel, texture) details as opposed to discriminative models requiring high-level (structural, object) details~\cite{DBLP:journals/corr/abs-2006-07733}.
However, there exist examples that show that they can help each other learn -- classifier guidance (discriminative) boosts the generative performance of Diffusion Models and StyleGANs~\cite{dhariwal2021diffusion, Sauer2021ARXIV} and generative augmentations help to enhance recognition capabilities~\cite{besnier2020dataset, zhang2021datasetgan, li2022bigdatasetgan, jahanian2022generative, sariyildiz2023fake}, etc. 
In this work, we discover that generative models, specifically pre-trained unconditional diffusion models, can not only help in discriminative tasks but inherently possess discriminative representation properties hidden inside.
These properties, which pre-trained diffusion models possess out-of-the-box, are intrinsically developed through their generative training process.

Recent works have also shown that generative models have similar recognition properties~\cite{Xiang_2023_ICCV, li2023your, Li_2023_ICCV, baranchuk2021label, tang2023dift, pnvr2023ld, xu2023openvocabulary, zhao2023unleashing}. 
The majority of such papers are based on diffusion models which are trained with text in addition to images~\cite{li2023your, tang2023dift, pnvr2023ld, xu2023openvocabulary, zhao2023unleashing}. 
However, high-quality captions are hard to obtain in many domains~\cite{yin2019automatic}.
In this work, we instead focus on investigating generative models as a self-supervised pretraining strategy~\cite{chen2020simple, caron2020unsupervised, DBLP:journals/corr/abs-2111-06377}, and including text obfuscates whether the diffusion helps performance or the text does, considering that text in the pretraining leaks the labels of the downstream task and defeats the purpose of being in the label-free unsupervised/self-supervised regime. 
Moreover, it has been seen that models trained with text losses are similar to supervised models~\cite{walmer2023teaching}. 
Besides, these works focus on limited tasks -- either only global (classification~\cite{Xiang_2023_ICCV, li2023your}) or only dense tasks (segmentation~\cite{Li_2023_ICCV, baranchuk2021label, pnvr2023ld, xu2023openvocabulary, zhao2023unleashing}, correspondence~\cite{tang2023dift, zhang2023tale}). 
In contrast, we do a comprehensive analysis of pre-trained text-free unconditional diffusion model features on a variety of downstream recognition tasks.

In particular, the selection of noise steps and feature blocks in the diffusion U-Net is not trivial and presents one of the main challenges with diffusion for discriminative representation.
We analyze diffusion features at various noise steps, feature blocks, and pooling combinations to understand which part of the model yields the most useful discriminative features and how they can be extracted.
We find that not only does performance drastically vary depending on these settings, but that there is also a substantial diversity of information among these features.
Additionally, the optimal configuration of these features varies by dataset.

To address these challenges, we propose a variety of feature extraction and aggregation mechanisms. Generally, classification heads extract information from a network's feature maps using a fixed pooling layer followed by a linear layer. But this choice is non-trivial and consequently, we propose a novel attention head for classification from diffusion features.
To unlock the full capabilities of diffusion models and incorporate the diversity among different blocks and noise steps, we propose two methods -- DifFormer and DifFeed, shown at a high level in Figure~\ref{fig:teaser_fig}.
With the first, we extend our attention mechanism as a feature fusion head, which we call DifFormer.
For the second, we propose a novel feedback mechanism for diffusion features, DifFeed, where we perform two forward passes on the diffusion model, the first as normal, and then the second where we perform learned combinations of U-Net decoder features at each encoder block.

We also investigate the performance of out-of-the-box diffusion features, DifFormer, and DifFeed, via benchmarking and analysis. 
We benchmark the unconditional model from guided diffusion (GD) (a.k.a. ablated diffusion model (ADM)) on a wide set of downstream tasks, \viz classification and semi-supervised classification on ImageNet~\cite{deng2009imagenet}, transfer learning for fine-grained visual classification (FGVC) on popular datasets~\cite{WahCUB_200_2011,maji13fine-grained, KrauseStarkDengFei-Fei_3DRR2013, KhoslaYaoJayadevaprakashFeiFei_FGVC2011, Nilsback08, Horn_2015_CVPR}, object detection/instance segmentation on COCO~\cite{lin2014microsoft}, and semantic segmentation on ADE20K~\cite{zhou2017scene}. 
Competitive performance on diverse tasks bolsters our hypothesis that unsupervised text-free diffusion features can serve as a unified representation.%

Given that diffusion models have quickly become ubiquitous in industry and academia, we hope that our insights on the capabilities of diffusion models could inspire future works that will achieve even better performance.
Additionally, we believe that our plug-and-play expansion mechanism of these pre-trained diffusion models to perform recognition will be of tremendous value to the community because one no longer needs to pre-train large, expensive models separately for different tasks. 
One can readily leverage the advancements in open-source generative foundational models for discriminative tasks.
Thus, we hope that our work's impact will grow as diffusion models become better, faster, and smaller.

In summary, our contributions are as follows:
\begin{itemize}
\item We analyze pre-trained unconditional diffusion features and discover that their discriminative power is distributed across network blocks, noise time steps, and feature resolutions. We observe that the best-performing feature configuration is dataset/task-dependent. 
\item We propose an attention mechanism for handling feature resolutions, and incorporate this attention mechanism as DifFormer, to combine the power of features from different blocks and noise steps, with less fixed pooling.
\item We propose a novel feedback module for diffusion, DifFeed, that modulates the features by feeding back the decoder features from the first forward pass to the encoder in the second pass for efficiency and better performance.
\item We demonstrate that diffusion models learn discriminative visual representations, not only for ImageNet classification but also FGVC, semi-supervised classification, object detection, and semantic segmentation.
\end{itemize}

\section{Related Work}
\label{sec:related_work}

\subsubsection{Generative Models.}
Generative models learn to sample novel images from a data distribution.
Generative Adversarial Networks (GANs)~\cite{goodfellow2020generative,karras2017progressive,brock2018large,Karras_2019_CVPR,karras2020analyzing,karras2020training,karras2021alias,Sauer2021ARXIV} are a class of such models that are trained by optimizing a min-max game between a generator, which synthesises images and a discriminator, which classifies the images as real or fake. 
Diffusion denoising probabilistic models (DDPM)~\cite{ho2020denoising}, a.k.a.\ diffusion models, are a class of likelihood-based generative models which learn a denoising Markov chain using variational inference. 
These models enjoy the benefit of having a likelihood-based objective like VAEs as well as high visual sample quality like GANs, even on high variability datasets. 
Diffusion models have proven to produce high-quality images~\cite{dhariwal2021diffusion}, beating previous SOTA generative models~\cite{brock2018large,razavi2019generating} 
for FID on ImageNet~\cite{deng2009imagenet}, and have also achieved amazing results in text-to-image generation~\cite{ramesh2022hierarchical,saharia2022photorealistic,rombach2022high}.
Application of these models is not just limited to generation but spans tasks like object detection~\cite{chen2022diffusiondet} and image segmentation~\cite{burgert2022peekaboo}.
We continue these lines of application with our comprehensive study and proposed methods.

\subsubsection{Discriminative Models.}
Discriminative models extract useful information from images that can then be used to solve downstream recognition tasks. 
Early self-supervised learning (SSL) methods attempt to learn image representations by training neural network backbones with partially degraded inputs to predict the missing information~\cite{zhang2016colorful,noroozi2016unsupervised,misra2020self,pathak2016context}.
More recently, many approaches revolve around a contrastive loss objective, min-maxing distance between positive and negative pairs~\cite{chen2020simple,chen2020big,DBLP:journals/corr/abs-2003-04297,chen2021mocov3,pmlr-v139-zbontar21a,tomasev2022pushing}. %
On the other hand, some methods operates without negative pairs~\cite{DBLP:journals/corr/abs-2006-07733,chen2021exploring,Bardes2021VICRegVR}, and others use clustering-style objectives~\cite{caron2018deep,caron2020unsupervised}.
MAE~\cite{DBLP:journals/corr/abs-2111-06377} and iBOT~\cite{zhou2022ibot} train an autoencoder via masked image modeling~\cite{assran2022masked,bao2022beit,huang2022contrastive}.
DINO~\cite{caron2021emerging} uses self-supervised knowledge distillation between various image views in Vision Transformers~\cite{dosovitskiy2020image}.
With all the recent advances, the latest SSL methods 
surpass supervised methods on many key discriminative baselines~\cite{pang2022unsupervised,oquab2023dinov2,zhou2022mugs,li2022efficient} but not generation.

\subsubsection{Unified Models.}
Unsupervised unified representation learning for generative and discriminative tasks, unsupervised unified representation learning in short, aims to learn general-purpose representations that can be used for various downstream tasks like image recognition, reconstruction, and synthesis~\cite{li2022mage,Xiang_2023_ICCV}.
Works like ~\cite{donahue2016adversarial,dumoulin2016adversarially,chen2016infogan,nie2020semi} leverage the unsupervised nature of GANs to learn good image representations.
BiGAN~\cite{donahue2016adversarial}, ALI~\cite{dumoulin2016adversarially}, ALAE~\cite{pidhorskyi2020adversarial}, and BigBiGAN~\cite{donahue2019large} do joint Encoder-Generator training with a discriminator on image-latent pairs. 
PatchVAE~\cite{Gupta_2020_CVPR} improves performance by encouraging the model to learn good mid-level patch representations.  
ViT-VQGAN~\cite{yu2021vector} learns to predict VQGAN tokens autoregressively in a rasterized fashion.
MAGE~\cite{li2022mage} uses a variable masking ratio during training and iterative decoding for inference to achieve both high-quality unconditional image generation and good classification results. It comes with the drawback of two heavy-weight trainings before training for downstream tasks. Moreover, its representations have only been shown to work for classification but not for localization or dense prediction tasks.   

\subsubsection{Diffusion Features.}
Recently, the use of diffusion models' intermediate activation features for various non-generative tasks is also gaining popularity. 
Works like ~\cite{baranchuk2021label,pnvr2023ld,Li_2023_ICCV, xu2023openvocabulary, zhao2023unleashing} focus on segmentation.
Diffusion Classifier~\cite{li2023your}, which solves zero-shot classification using Stable Diffusion (SD) noise prediction, also provides the performance of training a ResNet-50 using SD features as their baseline comparison for ImageNet classification (see appendix for our linear probing results with SD). 
DIFT~\cite{tang2023dift} employs SD features for semantic correspondence. 
DDAE~\cite{Xiang_2023_ICCV} also finds that using an unconditional diffusion model's features yields competitive results on CIFAR10~\cite{krizhevsky2009learning} and TinyImageNet~\cite{tiny-imagenet}. 
However, we observe that this competitiveness does not hold anymore for datasets with higher variability. Previous works are task specific and most of them use text-conditioned models which leads to leakage of labels into the downstream tasks.

\section{Analysis}
\label{sec:analysis}

\subsection{Preliminaries}

\subsubsection{Diffusion Models Fundamentals.}
In the forward noising process of diffusion models~\cite{ho2020denoising, nichol2021improved, dhariwal2021diffusion}, an image sample $x_0$ of the data distribution $q(x_0)$ is noised iteratively. Given a variance schedule ${\{\beta_t\}}_{t=1}^T$ (with $\alpha_t:=1-\beta_t$ and $\Bar{\alpha}_t:=\prod_{i=0}^t\alpha_i$), an intermediate noised state $x_t$ can be sampled using
\begin{equation}
\label{eq:make_noisy}
\begin{split}
x_t = \sqrt{\Bar{\alpha}_t}x_0 +\sqrt{1-\Bar{\alpha}_t}\epsilon, \epsilon \sim \mathcal{N}(0, \textbf{I})
\end{split}
\end{equation}
A neural network $\epsilon_\theta$ is used to reverse the diffusion process by learning to predict the noise that should be removed from $x_t$ to get $x_{t-1}$. Iterative denoising in this manner can be used to generate samples in the data distribution from pure Gaussian noise. Please find more details about diffusion models in the appendix.

\subsubsection{Diffusion Models Feature Extraction.}
In this work, we use the guided diffusion (GD) implementation, which uses a U-Net-style architecture with residual blocks for $\epsilon_\theta$. 
This implementation improves over the original \cite{ho2020denoising} architecture by adding multi-head self-attention at multiple resolutions, scale-shift norm, and using BigGAN~\cite{brock2018large}  residual blocks for upsampling and downsampling. 
We consider each of these residual blocks, residual+attention blocks, and downsampling/upsampling residual blocks as individual blocks and number them as $b \in \{1, 2, ..., 37\}$ (where $b=19$ corresponds to the output of the mid-block of the U-Net) for the pre-trained unconditional $256{\times}256$ guided diffusion model.

Our feature extraction is parameterized with the diffusion step $t$ and model block number $b$. 
We show an illustration of how input images vary at different time steps in Figure~\ref{fig:hypothesis_plot} bottom.
For feature extraction of image $x_0$, we use Eq. \ref{eq:make_noisy} to get noised image $x_t$. In the forward pass through the network $\epsilon_\theta(x_t,t)$, we use the activation after the block number $b$ as our feature vector $f_\theta(x_0,t,b)$.

\begin{table*}[t]
\begin{minipage}[t]{0.48\linewidth}

    \centering
    \includegraphics[width=1\linewidth]{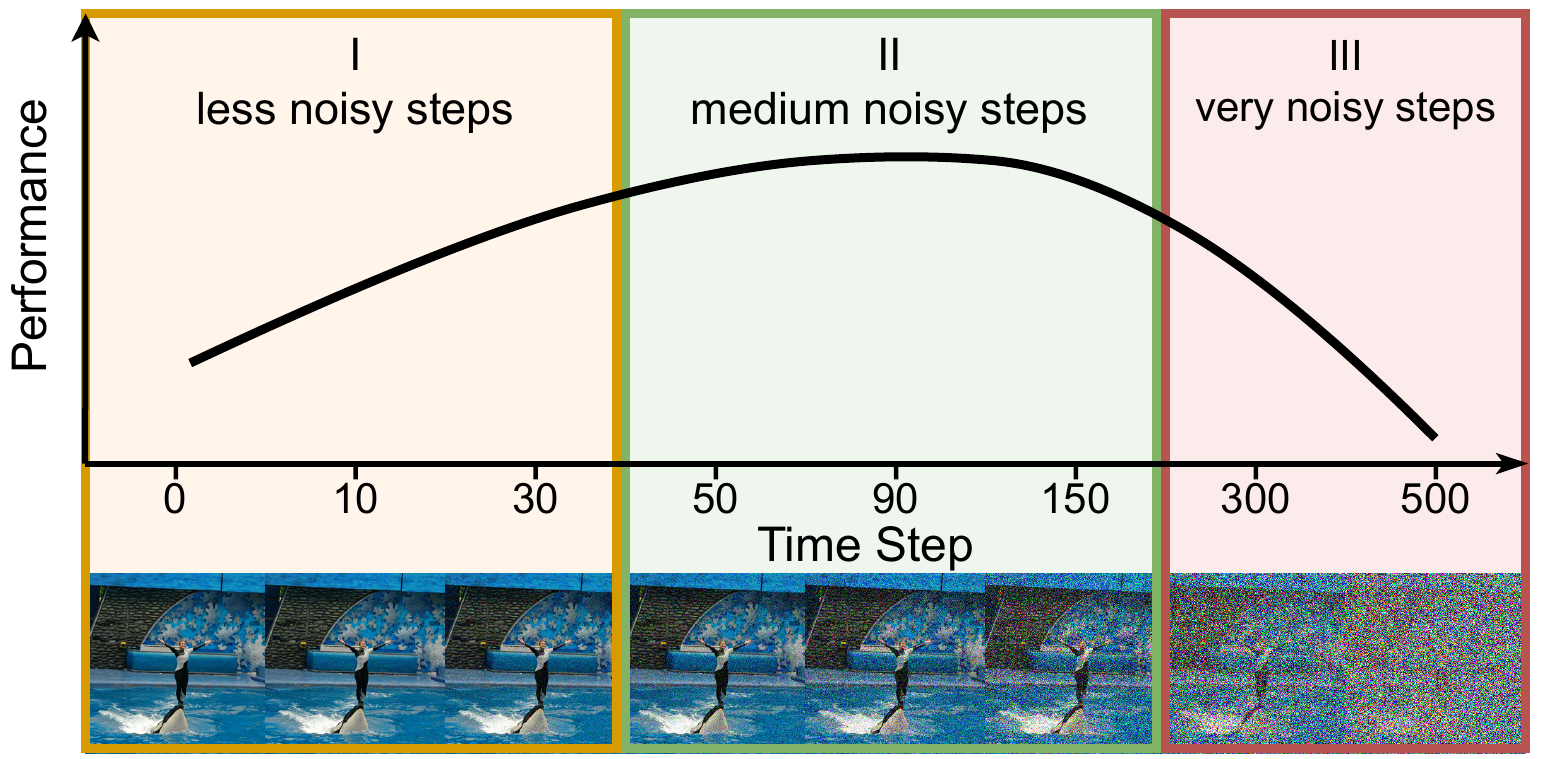}
    \captionof{figure}{\textbf{Hypothesis:} Diffusion features from low (region I) and high time step (region III) are not the most discriminative and have lower performance. The best features can be found in early-middle time steps (region II) and vary based on tasks/datasets. At low time steps, the diffusion model focuses more on stochastic details rather than structure, while at high time steps since the input is less recognizable, feature quality degrades.    }
    \label{fig:hypothesis_plot}
\end{minipage}
\hfill
\begin{minipage}[t]{0.48\linewidth}

    \centering
    \includegraphics[width=\linewidth]{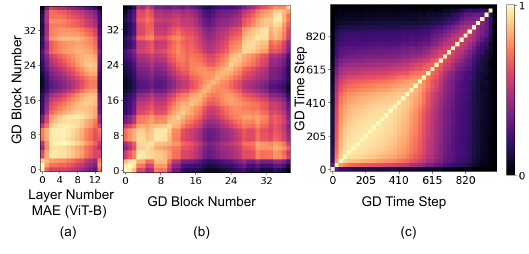}
    \captionof{figure}{Feature representation comparisons via centered kernel alignment (CKA). (a) Similarity of diffusion U-Net features across blocks at $t=90$ with features from MAE (ViT-B) layers. (b) Similarity across blocks of the diffusion U-Net at $t=90$. (c) Similarity across timesteps of features from U-Net block $b=24$. (a), (b), and (c) point toward the diffusion U-Net features being quite diverse. }
    \label{fig:cka_analysis}

\end{minipage}
\end{table*}

\subsection{Our Key Findings}

\begin{figure*}[t]
    \centering
    \includegraphics[width=\textwidth]{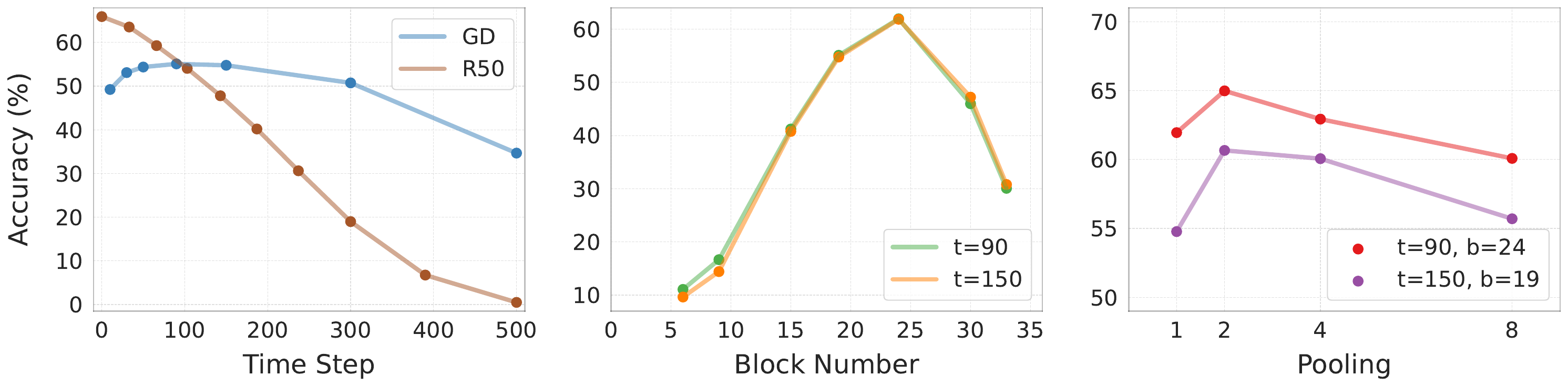}
    \caption{Ablations on ImageNet (1000 classes) with varying time steps, block numbers, and pooling size, for a linear classification head on frozen features. We find the model is least sensitive to pooling, and most sensitive to block number, although there is also a steep drop-off in performance as inputs and predictions become noisier. We further provide ResNet-50's (R50) performance over noisy time step images for comparison.}
    \label{fig:bn_t_acc}
\end{figure*}
\begin{figure*}[t]
    \centering
    \includegraphics[width=1.0\linewidth]{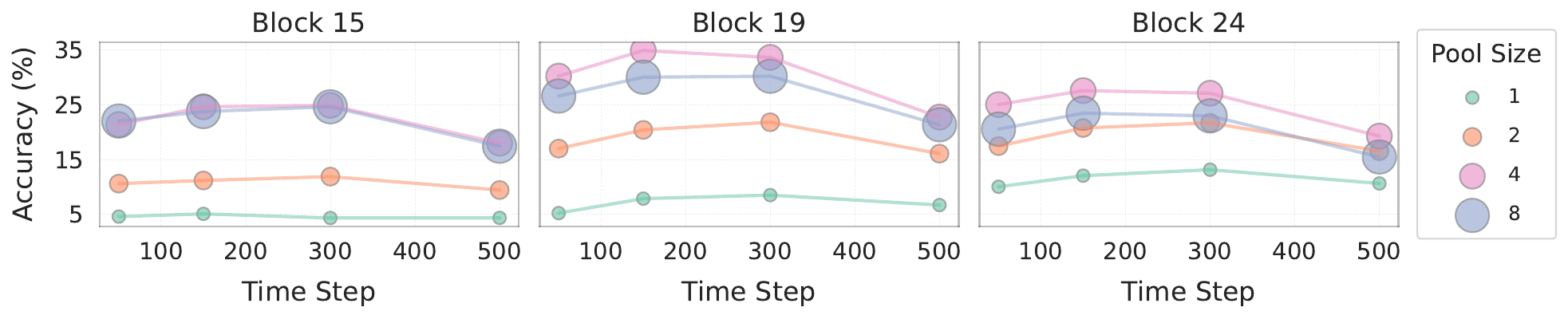}
    \caption{FGVC feature extraction analysis. We show accuracy for different block numbers, time steps, and pooling sizes. Block 19 is superior for FGVC, in contrast to ImageNet where 24 was ideal.}
    \label{fig:cub_search}
\end{figure*}

We hypothesize that early-middle noise steps and middle U-Net blocks may provide ideal features for classification in a setting where only a single block number and noise time step may be used (Figure~\ref{fig:hypothesis_plot}). Intuitively, when diffusion noise is too low, the noise prediction is too ``easy'' and the model does not need informative features to solve it (see Figure~\ref{fig:hypothesis_plot} bottom, $t{=}0$ vs. $10$); while when noise is too high, the discriminative portions of the image are themselves distorted (see $t{=}500$) leading to noisy prediction and less informative features. Hence, we hypothesize that there exists a sweet spot in the time step axis.
Existing literature using both pixel-space as well as latent diffusion models shares our intuition, especially with respect to time steps~\cite{li2023your,Xiang_2023_ICCV,Li_2023_ICCV,pnvr2023ld,tang2023dift}, and the only work that prefers inputs with no noise has low ablative granularity and does not try similar early-middle noise steps that work well in these other works~\cite{xu2023openvocabulary}.

First, we seek to verify that this phenomenon holds for larger images, for global prediction tasks.
Additionally, larger images naturally have larger feature maps, introducing an added axis which must be accounted for: pooling.
So, we explore the suitability of features for classification (by learning only a linear layer on top of a frozen U-Net backbone) on the axis of block number, noise time step, and feature pooling size, for images from larger, more realistic datasets, starting with ImageNet (training/inference details present in Section~\ref{sec:experiments}).
Since a dense grid search on ImageNet is quite resource-intensive, we first do a line search on time steps (roughly log-equidistant between $t{=}0$ and $500$) and find that $t{=}90{,}150$ perform best (Figure~\ref{fig:bn_t_acc}~(a)). 
Next, with these time steps, we do a line search on block number (chosen at even intervals, $b{=}19$ is bottleneck) to find $b{=}24{,}19$ to be the best (Figure~\ref{fig:bn_t_acc}~(b)). Finally, we use the best $b$ and $t$ settings for each pooling (Figure~\ref{fig:bn_t_acc}~(c)). 
This shows that, as expected, early noise steps and middle block numbers tend to work well.
Furthermore, pooling to larger features is not strictly better, as the best results for ImageNet use the second smallest features.

Additionally, we show a similar ablation for the Caltech Birds (CUB), a fine-grained visual classification (FGVC) task dataset,  
in Figure~\ref{fig:cub_search}.
Critically, we find a shift in optimal block numbers, time steps, and feature sizes, where some of these prefer earlier blocks, later time steps, and larger feature sizes.
Nevertheless, our claim in Figure \ref{fig:hypothesis_plot} still holds as a general principle.

Xiang \etal \cite{Xiang_2023_ICCV} provide some heuristics for choosing these, per-dataset, in a label-free manner; however, this still requires substantial intervention to adapt to each dataset.
Furthermore, the selection of a single block or time step is somewhat restrictive.
We thus investigate these representations further, this time in the form of Centered Kernel Alignment (CKA)~\cite{pmlr-v97-kornblith19a}, to judge, based on their similarities to each other, which set(s) we might use for varying tasks.

We use linear centered kernel alignment (CKA) to find the degree of similarity between the representations of different blocks of the diffusion model.
Following conventions from prior works~\cite{Gwilliam_2022_CVPR,walmer2023teaching}, we use the 2,500 image test set of ImageNet-50~\cite{vangansbeke2020scan}, a selection of 50 classes of ImageNet.
These results, shown in Figure~\ref{fig:cka_analysis}, provide clues that the representations are quite diverse.
With the block number and time step comparisons, we see that diffusion U-Net features differ greatly from each other depending on these crucial settings. (Interestingly, we see a peak in similarity among features close to the early-middle timesteps in Figure~\ref{fig:cka_analysis}~(c), supporting our hypothesis in Figure~\ref{fig:hypothesis_plot}). 
Furthermore, the comparison with ViT features, particularly comparing the last, most discriminative layer of the ViT with the layers near the U-Net bottleneck ($b=19$) suggests there is a variety of semantic information distributed across the features at varying blocks and noise steps.
We hypothesize that ensembling these features is the key for unlocking the power of diffusion models as representation learners.
Thus, we propose the feature aggregation methods detailed in the following sections.

\section{Our Proposed Feature Fusion}
\label{sec:approach}
One can have different strategies to explore feature aggregation.  
We start with a simple attention-based mechanism that extracts meaningful information from a feature vector. Next, we introduce two orthogonal but effective feature aggregation methods. First, we introduce a more involved version of the attention-based mechanism which combines features coming from multiple times steps and blocks. Second, we also introduce a feedback mechanism that modulates the feature extraction process, resulting in the efficient accumulation of features.
\subsection{Attention-based Classification Head}

\begin{figure*}[ht]
    \centering
    \includegraphics[width=0.95\linewidth]{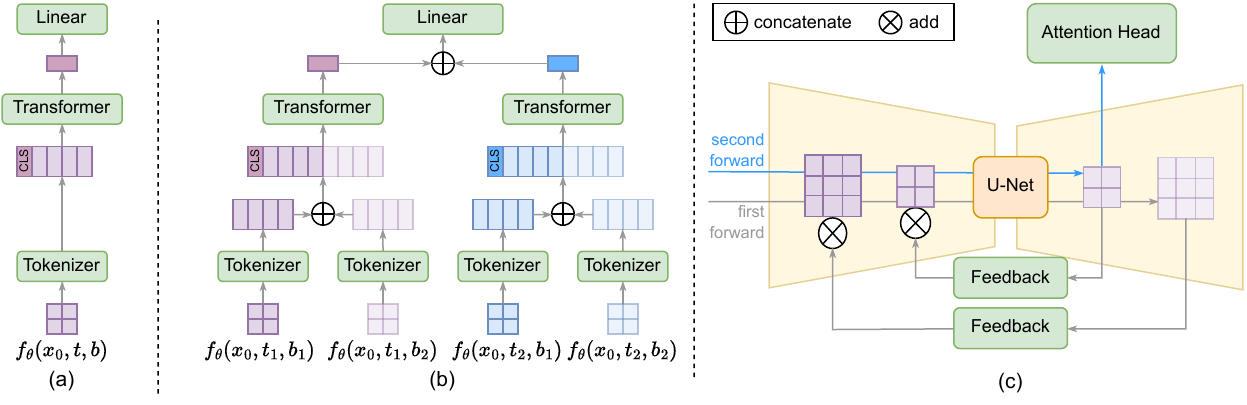}
    \caption{Architecture of 3 proposed methods for utilizing diffusion U-Net features. (a) \textbf{Attention head} takes a feature $f_\theta(x_0,t,b)$ of image $x_0$, diffusion time step $t$ and U-Net block number $b$, tokenizes it, adds a CLS token, passes it through a Transformer, and finally feeds the CLS token to a linear layer. (b) \textbf{DifFormer}, 
    similarly, 
    first tokenizes features from different blocks of the same timestep, concatenates them before feeding them to a Transformer.
    The CLS tokens from the Transformer outputs from various timesteps are concatenated and fed to a final linear layer. (Here, features of $\mathcal{T}=\{t_{1},t_{2}\}, \mathcal{B}=\{b_{1},b_{2}\}$ are selected.) (c) \textbf{DifFeed}, after the first forward pass extracts the decoder features, these are passed through a feedback network (a convolution layer with normalization and activation) and fed back into corresponding encoder blocks. In the second forward pass, the feedback is added to the encoder blocks to refine and fuse features from decoder layers into the encoder, finally to extract the features from a specific decoder block, \eg, $\mathcal{B}=\{b_{i}\}$, which is passed to an Attention head (a). }
    \label{fig:fusion_arc}
\end{figure*}

The two most common methods for evaluating the effectiveness of self-supervised pre-training for classification are linear probing and finetuning.
The first involves learning a linear head on a frozen backbone to predict a class label; the second is the same except the backbone is not frozen.
Since the diffusion U-Net is a convolutional architecture, we must use a combination of pooling and flattening to yield a feature vector for each image that serves as an input to the linear head.
However, as we note in Section~\ref{sec:analysis}, the selection of these pools is nontrivial.
So, we propose an Attention head to act as a learnable pooling mechanism.

Specifically, as shown in Figure~\ref{fig:fusion_arc} (a), we first use a tokenizer, consisting of adaptive average pooling, layer normalization, and $1 {\times} 1$ convolution, to reduce the selected feature map $f_{\theta}(x_{0},t,b)$ into $N_{b} {\times} N_{b} {\times} 1024$, where $N_{b}$ represents feature map size after pooling. We adopt a large pool size of 16 and only apply pooling when the feature map size is larger than this. We then flatten the feature map to generate $N_{b}^2$ tokens and append a CLS token before providing it as an input to our Transformer layer. We follow the standard Transformer architecture consisting of layer normalization and QKV-Attenton. Finally, the CLS token is extracted and used for classification by a linear layer.

\subsection{DifFormer: Transformer Feature Fusion}

To consolidate features $f_{\theta}(x_{0},t,b)$ from multiple times $t \in \mathcal{T} \subset \{1,2,\cdots, T\} $ and blocks $b \in \mathcal{B} \subset \{1, 2, ..., 37\}$, we propose an extension of Attention head (Figure~\ref{fig:fusion_arc} (b)). After the tokenizer, we concatenate all tokens from the same noise step, resulting in $ \sum_{b\in \mathcal{B}}N_{b}^2$ tokens for each time $t \in \mathcal{T}$. Each set is processed by a Transformer layer. Finally, all CLS tokens from multiple time steps are concatenated into a $1024 {\times} |\mathcal{T}|$ dimensional vector which is then input to a linear layer. Tokenizer and Transformer use the same parameters across all noise steps.

\subsection{DifFeed: Feedback Feature Fusion}

To enable faster fusion of features from various blocks, we propose a dynamic feedback mechanism shown in Figure~\ref{fig:fusion_arc} (c), similar to~\cite{shrivastava2016contextual}, which takes advantage of the U-Net architecture to refine diffusion features for the specific downstream task.
Specifically, we decouple the feature extraction process into two forward passes. 
In the first forward pass, we store the feature maps generated by the decoder at a selected set of blocks.
We then feed these features to a light-weight feedback network, which has unique layers consisting of $1 {\times} 1$ convolution, batch normalization, and ReLU for each selected decoder block.
These feedback layers learn to map decoder features to a suitable space for adding them to the corresponding encoder features. %
Our key intuition behind this design is that, as we find in Section~\ref{sec:analysis}, the decoder features contain important semantic information, unique to each block, that can be used to enrich the encoder image representations. 
To finally obtain the features for the downstream task, we perform a second forward pass and add the feedback features to the encoder features based on various feedback strategies (see Section \ref{subsec:ablations}). 
We use the attention head on top of one of the second forward pass features.%

\section{Experiments}
\label{sec:experiments}
In Section~\ref{subsec:main_results}, we compare our diffusion extraction to baselines and competing unified representation methods. %
We provide ablations in Section~\ref{subsec:ablations} to justify key design choices for Attention head, DifFormer, and DifFeed. 
We benchmark diffusion for key downstream tasks in Section~\ref{subsec:fgvc}, Section~\ref{subsec:semantic_seg}, and Section~\ref{subsec:object_det_seg}.

\begin{table*}[t]
\begin{minipage}{\linewidth}
\centering
\caption{
Main results. We compare self-supervised and unified learners in terms of classification and generation at resolution $256{\times}256$. Our aggregation mechanisms are unified learners post hoc, \ie, we use a pre-trained generative model for performing discrimination, while previous unified learners are jointly trained to do both generation and discrimination. We use \textbf{bold} to denote the best and \underline{underline} for the second-best.   
}
\label{tab:baselines}
\resizebox{0.8\textwidth}{!}{
\setlength{\tabcolsep}{10pt}
\begin{tabular}{@{}lcccc@{}}
\toprule
Method & Type & Training Strategy& Accuracy & FID  \\
\midrule
SimCLR$^\dagger$ &Self-Supervised& - &69.3\%&n/a\\
SwAV$^\ddagger$ &Self-Supervised& - &75.3\%&n/a\\
MAE$^\flat$ &Self-Supervised& - &73.5\%&n/a\\
\midrule
BigBiGAN* &Unified &jointly trained&   60.8\%	& 28.54 \\
MAGE$^\natural$  &Unified &jointly trained& \textbf{78.9\%}  & \textbf{9.10} \\
\midrule
U-Net Encoder& Supervised &-& 64.3\% & n/a \\
\midrule
GD  &Unified &post hoc& 64.9\%	& \underline{26.21}$^\S$ \\
Attention  &Unified &post hoc& 74.6\% & \underline{26.21}$^\S$ \\
DifFormer  &Unified &post hoc& 76.0\%  & \underline{26.21}$^\S$ \\
DifFeed  &Unified &post hoc&  \underline{77.0\%} & \underline{26.21}$^\S$ \\
\bottomrule
\end{tabular}
}

\hspace{-.02in}\footnotesize{$^\dagger$Result from \cite{chen2020simple}. $^\ddagger$Result from \cite{caron2020unsupervised}. $^\flat$Result from \cite{DBLP:journals/corr/abs-2111-06377}. 
$^\natural$Results from \cite{li2022mage}. $^\S$Results from \cite{dhariwal2021diffusion}. *Results from \cite{donahue2019large}. BigBiGAN's best FID is at generator resolution 128.}%
\end{minipage}
\end{table*}

\noindent\textbf{Experiment Details.} Unless otherwise specified, we use the unconditional ADM U-Net architecture from Guided Diffusion~\cite{dhariwal2021diffusion}  with total timesteps $T=1000$.
We use the $256{\times}256$ checkpoint; thus we resize all inputs to this size and use center-crop and flipping for data augmentation.
We use $f_{\theta}(x_{0},t=150,b=24)$ as a default feature map. 
We use cross entropy loss with an Adam optimizer~\cite{kingma2017adam} and follow the VISSL protocol for linear probing -- training for 28 epochs, with StepLR at $0.1$ gamma every 7 epochs.
However, we do not use random cropping or batch normalization.
Most of our experiments are run on 4 NVIDIA RTX A5000 GPUs.

\noindent\textbf{Datasets.}
The dataset we use for our main result is ImageNet-1k~\cite{deng2009imagenet} which contains 1.3M training images over 1000 classes.
Additionally, we run ablations and similar explorations on Caltech-UCSD (CUB)~\cite{WahCUB_200_2011} and ImageNet-1k with 1\% label (IN-1\%). CUB is a fine-grained visual classification dataset consisting of 200 classes. IN-1\% uses all the training images of ImageNet-1k but only 1\% of the labels.   
Please see the appendix for details.

\subsection{Main Results: ImageNet Classification}
\label{subsec:main_results}
First, we show the linear probing performance of diffusion in Table~\ref{tab:baselines}. For guided diffusion with linear head (GD), we use a $2{\times}2$ adaptive average pooling to reduce the spatial dimension.
As a baseline, we compare to the pre-trained guided diffusion classifier (used for classfier-guidance~\cite{dhariwal2021diffusion}), since it uses the same U-Net Encoder.
We also offer a comparison to other unified models: BigBiGAN~\cite{donahue2019large} and MAGE~\cite{li2022mage} as well as self-supervised models like SimCLR~\cite{chen2020simple} and SwAV~\cite{caron2020unsupervised}. 
GD outperforms BigBiGAN in terms of both generation and classification, especially when BigBiGAN is forced to handle the higher resolution, $256{\times}256$ images.
Hence, diffusion models beat GANs for image classification (and generation). This is not yet comparable to classification-only models or the powerful unified MAGE model, with a gap of over $10\%$ top-1 accuracy.

As described previously, we propose several approaches to deal with the large spatial and channel dimensions of U-Net representations as well as to fuse features from various blocks. For these proposals, we use settings selected via the ablations described in Section~\ref{subsec:ablations}.
In Table~\ref{tab:baselines}, we try the best-performing Attention head on ImageNet-1k and find it significantly outperforms linear probe. %
This suggests the classification head is an important mechanism for extracting useful representations from diffusion models, and it could be extended to other generative models.
Similarly, using better feature fusion mechanisms like DifFormer and DifFeed, with frozen backbones, can lead to significant improvement in performance to the level that it is comparable to MAGE, while better than SimCLR and SwAV (see Table~\ref{tab:baselines}). Compared to our light-weight aggregation strategies trained on top of fixed pretrained GD backbone, MAGE is a resource-heavy pipeline that first trains a VQGAN (or uses a pre-trained VQGAN tokenizer), then learns masked token modeling (which requires 4 days on 32 NVIDIA V100 GPUs), and then does downstream training.

\subsubsection{Ablations.}
\label{subsec:ablations}

\begin{table*}[!t]
\begin{minipage}[t]{0.4\linewidth}
\centering
\caption{
Attention Head Ablation. We compare the effect of the number of layers on top-1 accuracy and on the number of Attention head network parameters. We freeze the U-Net and train the Attention heads for 15 epochs.  
}
\label{tab:ablation-attention}
\resizebox{\linewidth}{!}{
\setlength{\tabcolsep}{10pt}
\begin{tabular}{@{}c c c c@{}}
\toprule
 \multirow{2}{*}{\# Layers}& \multicolumn{2}{@{}c@{}}{Accuracy} & \multirow{2}{*}{\# Params}   \\
 \cmidrule{2-3}

 & IN-1\%  & CUB &  \\
\midrule

1 & 40.4\% & 39.4\% & 15.2M \\
2 & 45.7\% & 47.4\% & 27.8M \\
4 & \textbf{47.9\%} & \textbf{52.0\%} & 53.0M \\

\bottomrule
\end{tabular}
}

\end{minipage}
\hfill
\begin{minipage}[t]{0.56\linewidth}
\centering
\caption{
DifFormer Ablation. We compare various combinations of time and block fusion strategies on top-1 accuracy and the number of forward passes required through the U-Net backbone. \# of forwards is the key driver of the relative time cost differences during training and inference.
}
\label{tab:ablation-fusion}
\resizebox{\textwidth}{!}{
\setlength{\tabcolsep}{10pt}
\begin{tabular}{@{}c cc c c@{}}
\toprule
 \multirow{2}{*}{$\mathcal{T}$} &\multirow{2}{*}{$\mathcal{B}$} &\multicolumn{2}{c}{Accuracy} & \multirow{2}{*}{\# Forwards}   \\
 \cmidrule(lr){3-4}

 &  & IN-1\%  & CUB &  \\
\midrule

$\{150\}$ &  $\{24\}$ 	&45.7\% & 47.4\%  & 1 \\
$\{90,150,300\}$ &   $\{24\}$	&49.0\%& 54.1\% & 3 \\
$\{150\}$ & $\{19,24,30\}$   	&43.2\%& 48.8\%  & 1 \\
$\{90,150,300\}$ & $\{19,24,30\}$   	&\textbf{50.4\%}& \textbf{56.8\%}  & 3\\
\bottomrule
\end{tabular}
}
\end{minipage}
\end{table*}
\label{subsec:semantic_seg}

\begin{table*}[!t]
\begin{minipage}[t]{0.57\linewidth}
\centering
\caption{
DifFeed Ablation. We compare various feedback block sampling strategies on top-1 accuracy, on the number of forward passes required through the U-Net backbone, and on the number of feedback network parameters.
}
\label{tab:ablation-feedback}
\resizebox{\linewidth}{!}{
\setlength{\tabcolsep}{10pt}
\begin{tabular}{@{}l c c c c@{}}
\toprule
 \multirow{2}{*}{Block Sampling}  &\multicolumn{2}{c}{Accuracy} & \multirow{2}{*}{\# Forwards} & \multirow{2}{*}{\# Params}   \\
 \cmidrule(lr){2-3}

 &   IN-1\%  & CUB & & \\
\midrule

all  	&31.1\% & 57.2\%  & 2 & 7.6M\\
bottleneck    	& \textbf{51.5}\% & \textbf{71.0}\%  & 2 & 2.8M  \\
windowed 	&49.4\%& 67.1\% & 2 & 5.2M\\
multi-scale   	&49.4\%& 66.0\%  & 2 & 5.2M \\
\bottomrule
\end{tabular}
}

\end{minipage}
\hfill
\begin{minipage}[t]{0.39\linewidth}
\centering
\caption{Semi-supervised results. We give accuracy results on ImageNet-1k with 1\% labels and 10\% labels for fine-tuning.}
\label{tab:semi-supervised}
    \resizebox{0.8\textwidth}{!}{
    \setlength{\tabcolsep}{10pt}
    \begin{tabular}{@{}l c c@{}} 
        \toprule
        Method & 1\% label  & 10\% label   \\
        \midrule
        SimCLR$^\dagger$   &  48.3\%  & 65.6\% \\ %
        SwAV$^\ddagger$     &  \textbf{53.9}\%  & \textbf{70.2}\%  \\ %
        \midrule
        GD  &  46.7\% & 64.4\%         \\
        DifFormer  & 51.8\%  &    66.2\%   \\ %
        DifFeed &  \underline{52.9 \%} &  \underline{66.6\%}   \\ %
        \bottomrule
    \end{tabular}
}

\hspace{-.02in}\footnotesize{
$^\dagger$From \cite{chen2020simple}. $^\ddagger$From \cite{caron2020unsupervised}.}%

\end{minipage}
\end{table*}

Here, we give the results of the ablation study and show the best settings for our Attention head, DifFormer, and DifFeed. Note that all ablation results are obtained with frozen backbones. From Table~\ref{tab:ablation-attention}, we find that using multiple Transformer layers in Attention head leads to better accuracy. Considering the increased number of parameters, we adopt two Transformer layers in our Attention head. Table ~\ref{tab:ablation-fusion} compares the effect of various combinations of time and block in DifFormer. We find that noise step fusion substantially improves the accuracy, and combining it with block fusion increases the accuracy further. Thus we adopt $\mathcal{T}=\{90,150,300\}$ and  $\mathcal{B}=\{19,24,30\}$ in DifFormer. 

In Table ~\ref{tab:ablation-feedback}, we examine four block selection strategies for generating feedback features that can be fed to the encoder layers. For all experiments, we use $t=150$ and set $b=24$ for extracting the final feature. The naive ``all'' strategy uses all decoder blocks to generate feedback features and maps them to symmetrically corresponding encoder blocks. 
The rest of the 
strategies 
aim to find an optimal sparser selection of blocks. 
Our ``bottleneck'' strategy uses only the bottleneck blocks $\mathcal{B}=\{21,24,27,30,33,36\}$ while ``windowed'' strategy only generates feedback from the first 5 blocks of the decoder. Our ``multi-scale'' strategy uses the same blocks as ``windowed'' strategy but feeds them back to a singular block in the encoder. 
We adopt ``bottleneck'' strategy for DifFeed as it works the best.

\subsection{Semi-Supervised Classification}
\label{subsec:semi-sup}

In this subsection, we evaluate the power of our proposed methods on datasets with limited labels. Following SimCLR~\cite{chen2020simple}, we sample 1\% and 10\% of the labeled ImageNet-1k in a class-balanced way and call them IN-1\% and IN-10\% respectively. Table \ref{tab:semi-supervised} shows the results of SimLR, SwAV, and our proposed models, with fine-tuned backbones. We observe that GD has worse performance compared to SimCLR but using DifFeed and DifFeed for feature fusion improves the performance both on 1\% and 10\%. Ultimately, DifFeed's top-1 accuracy on the ImageNet validation set is better than SimCLR and slightly worse than SwAV.      

\begin{figure*}[t]
    \centering
    \includegraphics[width=1.0\linewidth]{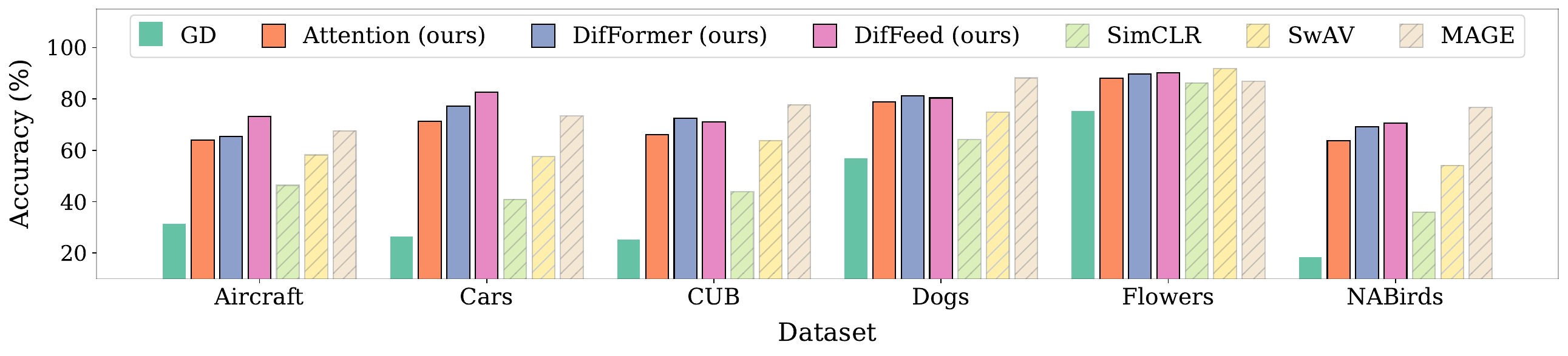}
    \caption{Fine-Grained Visual Classification (FGVC) results. We compare the diffusion baseline, GD, to our Attention head, DifFormer, and DifFeed in terms of accuracy for these popular FGVC datasets. We show that our methods are typically better than SimCLR and SwAV, and even better than MAGE for Aircraft, Cars, and NABirds.}
    \label{fig:fgvc}
\end{figure*}

\subsection{Fine-grained Visual Classification (FGVC)}
\label{subsec:fgvc}
Here, we give results for applying our method in the transfer setting for various FGVC datasets. 
We use both standard linear probing, as well as our proposed Attention head, DifFormer, and DifFeed with frozen backbones. %
We show these results in Figure~\ref{fig:fgvc}. We observe a gap in performance between GD and SimCLR.
However, all our methods - Attention head, DifFormer, and DifFeed, cover this performance gap and outperform self-supervised methods like SimCLR and SwAV on all datasets except Flowers, and beat MAGE on half of the datasets.

\subsection{Semantic Segmentation}

\begin{table*}[!t]
\begin{minipage}[t]{0.43\linewidth}
\centering
\caption{Semantic segmentation results on ADE20K~\cite{zhou2017scene} using UperNet~\cite{xiao2018unified} finetuned for 16K iterations. We report mIoU at a single scale.}%
\label{tab:semantic-segmentation}
    \resizebox{.9\textwidth}{!}{
    \setlength{\tabcolsep}{10pt}
    \begin{tabular}{@{}l  c@{}} 
        \toprule
        Method & mIoU \\
        \midrule
        Supervised & 40.9\%$^\sharp$ \\
        SimCLR & 39.9\%$^\sharp$ \\
        SwAV & 41.2\%$^\sharp$ \\
        DreamTeacher & 42.5\%$^\sharp$ \\ 
        \midrule
        MAE (ViT-B)  & 40.8\% \\
        MAE (ViT-L)  & \textbf{45.8}\% \\
        \midrule
        GD ($t=90, b=24$) & 42.3\% \\
        GD ($t=90, \mathcal{B}=\{33,30,27,24\}$) & \underline{44.3\%} \\
        DifFeed & 44.0\% \\
        \bottomrule
    \end{tabular}
}

\hspace{-.02in}\footnotesize{$^\sharp$ Results from \cite{Li_2023_ICCV}.}%
\end{minipage}
\hfill
\begin{minipage}[t]{0.53\linewidth}
    \centering
        \captionof{figure}{Detection results on COCO. Diffusion outperforms the supervised ResNet-50~\cite{he2015deep} Mask R-CNN~\cite{he2018mask} in terms of both object detection and instance segmentation, but is surpassed by the MAE-trained ViT-L.}
    \includegraphics[width=\linewidth]{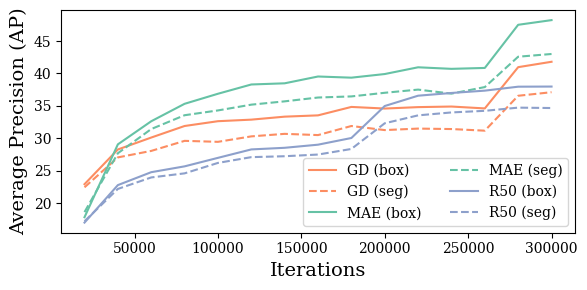}

    \label{fig:detection}
\end{minipage}
\end{table*}
\label{subsec:semantic_seg}

To evaluate GD features and DifFeed on dense prediction task we choose ADE20K semantic segmentation using UperNet~\cite{xiao2018unified} following MAE~\cite{DBLP:journals/corr/abs-2111-06377}. Note that the UperNet setup already has a 
multi-resolution feature fusion component 
and hence we don't use DifFormer here. We use batch size 2 (instead of 16) due to resource constraints in all the models we train. In Table \ref{tab:semantic-segmentation}, we observe that using only one block of GD for all UperNet inputs performs comparably to DreamTeacher~\cite{Li_2023_ICCV} (which uses distilled GD features), while better than ResNet-50 based model supervised, SimCLR~\cite{chen2020simple}, and SwAV~\cite{caron2020unsupervised}. 
Using multiple GD decoder blocks yields 2\% boost.
We observe a similar boost with DifFeed, where we extract features from $\mathcal{B}=\{33,30,27,24\}$ at $t=90$ for UperNet inputs. Even though these performances are better than MAE (ViT-B), MAE (ViT-L) has a superior result. More training details are present in the appendix.

\subsection{Object Detection and Segmentation}
\label{subsec:object_det_seg}

To continue our benchmark of diffusion model features, we evaluate them for object detection and segmentation.
Specifically, borrowing the settings of \cite{DBLP:journals/corr/abs-2111-06377}, we compare GD features taken fron time step $t = 75$ and block number $b = 24$, to MAE ViT-L~\cite{dosovitskiy2020image} and supervised ResNet-50~\cite{he2015deep} for AP on COCO in Figure~\ref{fig:detection}.
While worse than MAE, this is still a promising result, since with only due diligence hyperparameter tuning, it is already able to outperform the ResNet-based MaskR-CNN.
Due to the overall expense of training for these tasks, we leave further explorations to future work, and strongly encourage the community to examine this direction further as a promising application for diffusion as unified representation.
For more details on settings, refer to the appendix.

\section{Conclusion}

\noindent\textbf{Limitations.} As mentioned in Section~\ref{sec:intro}, this work is not focused on achieving or beating SOTA, although as we explore in Section~\ref{sec:experiments}, we perform quite well on many tasks, and outperform all competing methods on at least some dataset, and perform a broader variety of tasks than the methods we directly compare to.
Nevertheless, in some of the tasks, even with our plug-and-play improvements, diffusion does not give the best performance. But we believe that with proper hyper-parameter tuning and architectural changes such a feat can be achieved.
Diffusion models are also large and slow compared to some other networks (especially smaller CNNs).
Future work could focus on addressing these issues.

\noindent\textbf{Broader Impacts.} Recognition systems, both dense and global, have many good applications. They can power self-driving cars, help with tracking endangered species, improve early cancer detection, and more. 
This is especially true for models that train without supervision -- for conservation and medical domains, acquiring high quality labels is much more difficult than raw visual data.
However, such systems can also be used for surveillance and warfare. Since our work is focused on improving recognition, it could have impact in these areas.

In this paper, we present an approach for using the representations learned by diffusion models for recognition tasks.
This re-positions diffusion models as potential state-of-the-art unified self-supervised representation learners. 
We propose the novel DifFormer and DifFeed methods for learned feature extraction, pooling, and synthesis.
We demonstrate promising results for ImageNet classification, semi-supervised classification, FGVC transfer learning, object detection, and semantic segmentation.
Finally, to answer our titular question -- yes, text-free diffusion models do learn discriminative visual features and  
we suggest that with such promising out-of-the-box capabilities, diffusion is a prime candidate for unified representation learning, and with our initial benchmark results, we hope to encourage others to explore these directions.

\noindent\textbf{Acknowledgements.} We thank Pulkit Kumar for his assistance with generating figures and polishing the manuscript. This work was partially supported by NSF CAREER Award (\#2238769) to Abhinav Shrivastava. The authors acknowledge UMD’s supercomputing resources made available for conducting this research. The U.S. Government is authorized to reproduce and distribute reprints for Governmental purposes notwithstanding any copyright annotation thereon. Disclaimer: The views and conclusions contained herein are those of the authors and should not be interpreted as necessarily representing the official policies or endorsements, either expressed or implied, of the NSF or the U.S. Government.

% \bibliography{main}

\begin{thebibliography}{10}
\providecommand{\url}[1]{\texttt{#1}}
\providecommand{\urlprefix}{URL }
\providecommand{\doi}[1]{https://doi.org/#1}

\bibitem{assran2022masked}
Assran, M., Caron, M., Misra, I., Bojanowski, P., Bordes, F., Vincent, P.,
  Joulin, A., Rabbat, M., Ballas, N.: Masked siamese networks for
  label-efficient learning (2022)

\bibitem{bao2022beit}
Bao, H., Dong, L., Piao, S., Wei, F.: Beit: Bert pre-training of image
  transformers (2022)

\bibitem{baranchuk2021label}
Baranchuk, D., Voynov, A., Rubachev, I., Khrulkov, V., Babenko, A.:
  Label-efficient semantic segmentation with diffusion models. In:
  International Conference on Learning Representations (2021)

\bibitem{Bardes2021VICRegVR}
Bardes, A., Ponce, J., LeCun, Y.: Vicreg: Variance-invariance-covariance
  regularization for self-supervised learning. ArXiv  \textbf{abs/2105.04906}
  (2021)

\bibitem{besnier2020dataset}
Besnier, V., Jain, H., Bursuc, A., Cord, M., P{\'e}rez, P.: This dataset does
  not exist: training models from generated images. In: ICASSP 2020-2020 IEEE
  International Conference on Acoustics, Speech and Signal Processing (ICASSP).
  pp.~1--5. IEEE (2020)

\bibitem{brock2018large}
Brock, A., Donahue, J., Simonyan, K.: Large scale gan training for high
  fidelity natural image synthesis. arXiv preprint arXiv:1809.11096  (2018)

\bibitem{burgert2022peekaboo}
Burgert, R., Ranasinghe, K., Li, X., Ryoo, M.S.: Peekaboo: Text to image
  diffusion models are zero-shot segmentors. arXiv preprint arXiv:2211.13224
  (2022)

\bibitem{caron2018deep}
Caron, M., Bojanowski, P., Joulin, A., Douze, M.: Deep clustering for
  unsupervised learning of visual features. In: Proceedings of the European
  Conference on Computer Vision (ECCV). pp. 132--149 (2018)

\bibitem{caron2020unsupervised}
Caron, M., Misra, I., Mairal, J., Goyal, P., Bojanowski, P., Joulin, A.:
  Unsupervised learning of visual features by contrasting cluster assignments.
  Advances in Neural Information Processing Systems  \textbf{33},  9912--9924
  (2020)

\bibitem{caron2021emerging}
Caron, M., Touvron, H., Misra, I., J\'egou, H., Mairal, J., Bojanowski, P.,
  Joulin, A.: Emerging properties in self-supervised vision transformers. In:
  Proceedings of the International Conference on Computer Vision (ICCV) (2021)

\bibitem{chen2022diffusiondet}
Chen, S., Sun, P., Song, Y., Luo, P.: Diffusiondet: Diffusion model for object
  detection. arXiv preprint arXiv:2211.09788  (2022)

\bibitem{chen2020simple}
Chen, T., Kornblith, S., Norouzi, M., Hinton, G.: A simple framework for
  contrastive learning of visual representations. In: International conference
  on machine learning. pp. 1597--1607. PMLR (2020)

\bibitem{chen2020big}
Chen, T., Kornblith, S., Swersky, K., Norouzi, M., Hinton, G.E.: Big
  self-supervised models are strong semi-supervised learners. Advances in
  neural information processing systems  \textbf{33},  22243--22255 (2020)

\bibitem{chen2016infogan}
Chen, X., Duan, Y., Houthooft, R., Schulman, J., Sutskever, I., Abbeel, P.:
  Infogan: Interpretable representation learning by information maximizing
  generative adversarial nets. Advances in neural information processing
  systems  \textbf{29} (2016)

\bibitem{DBLP:journals/corr/abs-2003-04297}
Chen, X., Fan, H., Girshick, R.B., He, K.: Improved baselines with momentum
  contrastive learning. CoRR  \textbf{abs/2003.04297} (2020),
  \url{https://arxiv.org/abs/2003.04297}

\bibitem{chen2021exploring}
Chen, X., He, K.: Exploring simple siamese representation learning. In:
  Proceedings of the IEEE/CVF conference on computer vision and pattern
  recognition. pp. 15750--15758 (2021)

\bibitem{chen2021mocov3}
Chen*, X., Xie*, S., He, K.: An empirical study of training self-supervised
  vision transformers. arXiv preprint arXiv:2104.02057  (2021)

\bibitem{deng2009imagenet}
Deng, J., Dong, W., Socher, R., Li, L.J., Li, K., Fei-Fei, L.: Imagenet: A
  large-scale hierarchical image database. In: 2009 IEEE conference on computer
  vision and pattern recognition. pp. 248--255. Ieee (2009)

\bibitem{dhariwal2021diffusion}
Dhariwal, P., Nichol, A.: Diffusion models beat gans on image synthesis (2021)

\bibitem{donahue2016adversarial}
Donahue, J., Kr{\"a}henb{\"u}hl, P., Darrell, T.: Adversarial feature learning.
  arXiv preprint arXiv:1605.09782  (2016)

\bibitem{donahue2019large}
Donahue, J., Simonyan, K.: Large scale adversarial representation learning.
  Advances in neural information processing systems  \textbf{32} (2019)

\bibitem{dosovitskiy2020image}
Dosovitskiy, A., Beyer, L., Kolesnikov, A., Weissenborn, D., Zhai, X.,
  Unterthiner, T., Dehghani, M., Minderer, M., Heigold, G., Gelly, S., et~al.:
  An image is worth 16x16 words: Transformers for image recognition at scale.
  arXiv preprint arXiv:2010.11929  (2020)

\bibitem{dumoulin2016adversarially}
Dumoulin, V., Belghazi, I., Poole, B., Mastropietro, O., Lamb, A., Arjovsky,
  M., Courville, A.: Adversarially learned inference. arXiv preprint
  arXiv:1606.00704  (2016)

\bibitem{goodfellow2020generative}
Goodfellow, I., Pouget-Abadie, J., Mirza, M., Xu, B., Warde-Farley, D., Ozair,
  S., Courville, A., Bengio, Y.: Generative adversarial networks.
  Communications of the ACM  \textbf{63}(11),  139--144 (2020)

\bibitem{DBLP:journals/corr/abs-2006-07733}
Grill, J., Strub, F., Altch{\'{e}}, F., Tallec, C., Richemond, P.H.,
  Buchatskaya, E., Doersch, C., Pires, B.{\'{A}}., Guo, Z.D., Azar, M.G., Piot,
  B., Kavukcuoglu, K., Munos, R., Valko, M.: Bootstrap your own latent: {A} new
  approach to self-supervised learning. CoRR  \textbf{abs/2006.07733} (2020),
  \url{https://arxiv.org/abs/2006.07733}

\bibitem{Gupta_2020_CVPR}
Gupta, K., Singh, S., Shrivastava, A.: Patchvae: Learning local latent codes
  for recognition. In: Proceedings of the IEEE/CVF Conference on Computer
  Vision and Pattern Recognition (CVPR) (June 2020)

\bibitem{Gwilliam_2022_CVPR}
Gwilliam, M., Shrivastava, A.: Beyond supervised vs. unsupervised:
  Representative benchmarking and analysis of image representation learning.
  In: Proceedings of the IEEE/CVF Conference on Computer Vision and Pattern
  Recognition (CVPR). pp. 9642--9652 (June 2022)

\bibitem{DBLP:journals/corr/abs-2111-06377}
He, K., Chen, X., Xie, S., Li, Y., Doll{\'{a}}r, P., Girshick, R.B.: Masked
  autoencoders are scalable vision learners. CoRR  \textbf{abs/2111.06377}
  (2021), \url{https://arxiv.org/abs/2111.06377}

\bibitem{he2018mask}
He, K., Gkioxari, G., Dollár, P., Girshick, R.: Mask r-cnn (2018)

\bibitem{he2015deep}
He, K., Zhang, X., Ren, S., Sun, J.: Deep residual learning for image
  recognition (2015)

\bibitem{ho2020denoising}
Ho, J., Jain, A., Abbeel, P.: Denoising diffusion probabilistic models.
  Advances in Neural Information Processing Systems  \textbf{33},  6840--6851
  (2020)

\bibitem{huang2022contrastive}
Huang, Z., Jin, X., Lu, C., Hou, Q., Cheng, M.M., Fu, D., Shen, X., Feng, J.:
  Contrastive masked autoencoders are stronger vision learners (2022)

\bibitem{jahanian2022generative}
Jahanian, A., Puig, X., Tian, Y., Isola, P.: Generative models as a data source
  for multiview representation learning. In: International Conference on
  Learning Representations (2022),
  \url{https://openreview.net/forum?id=qhAeZjs7dCL}

\bibitem{karras2017progressive}
Karras, T., Aila, T., Laine, S., Lehtinen, J.: Progressive growing of gans for
  improved quality, stability, and variation. arXiv preprint arXiv:1710.10196
  (2017)

\bibitem{karras2020training}
Karras, T., Aittala, M., Hellsten, J., Laine, S., Lehtinen, J., Aila, T.:
  Training generative adversarial networks with limited data. Advances in
  Neural Information Processing Systems  \textbf{33},  12104--12114 (2020)

\bibitem{karras2021alias}
Karras, T., Aittala, M., Laine, S., H{\"a}rk{\"o}nen, E., Hellsten, J.,
  Lehtinen, J., Aila, T.: Alias-free generative adversarial networks. Advances
  in Neural Information Processing Systems  \textbf{34},  852--863 (2021)

\bibitem{Karras_2019_CVPR}
Karras, T., Laine, S., Aila, T.: A style-based generator architecture for
  generative adversarial networks. In: Proceedings of the IEEE/CVF Conference
  on Computer Vision and Pattern Recognition (CVPR) (June 2019)

\bibitem{karras2020analyzing}
Karras, T., Laine, S., Aittala, M., Hellsten, J., Lehtinen, J., Aila, T.:
  Analyzing and improving the image quality of stylegan. In: Proceedings of the
  IEEE/CVF conference on computer vision and pattern recognition. pp.
  8110--8119 (2020)

\bibitem{KhoslaYaoJayadevaprakashFeiFei_FGVC2011}
Khosla, A., Jayadevaprakash, N., Yao, B., Fei-Fei, L.: Novel dataset for
  fine-grained image categorization. In: First Workshop on Fine-Grained Visual
  Categorization, IEEE Conference on Computer Vision and Pattern Recognition.
  Colorado Springs, CO (June 2011)

\bibitem{kingma2017adam}
Kingma, D.P., Ba, J.: Adam: A method for stochastic optimization (2017)

\bibitem{pmlr-v97-kornblith19a}
Kornblith, S., Norouzi, M., Lee, H., Hinton, G.: Similarity of neural network
  representations revisited. In: Chaudhuri, K., Salakhutdinov, R. (eds.)
  Proceedings of the 36th International Conference on Machine Learning.
  Proceedings of Machine Learning Research, vol.~97, pp. 3519--3529. PMLR
  (09--15 Jun 2019), \url{https://proceedings.mlr.press/v97/kornblith19a.html}

\bibitem{KrauseStarkDengFei-Fei_3DRR2013}
Krause, J., Stark, M., Deng, J., Fei-Fei, L.: 3d object representations for
  fine-grained categorization. In: 4th International IEEE Workshop on 3D
  Representation and Recognition (3dRR-13). Sydney, Australia (2013)

\bibitem{krizhevsky2009learning}
Krizhevsky, A., Hinton, G., et~al.: Learning multiple layers of features from
  tiny images  (2009)

\bibitem{li2023your}
Li, A.C., Prabhudesai, M., Duggal, S., Brown, E.L., Pathak, D.: Your diffusion
  model is secretly a zero-shot classifier. In: ICML 2023 Workshop on
  Structured Probabilistic Inference {\&} Generative Modeling (2023),
  \url{https://openreview.net/forum?id=Ck3yXRdQXD}

\bibitem{li2022efficient}
Li, C., Yang, J., Zhang, P., Gao, M., Xiao, B., Dai, X., Yuan, L., Gao, J.:
  Efficient self-supervised vision transformers for representation learning
  (2022)

\bibitem{Li_2023_ICCV}
Li, D., Ling, H., Kar, A., Acuna, D., Kim, S.W., Kreis, K., Torralba, A.,
  Fidler, S.: Dreamteacher: Pretraining image backbones with deep generative
  models. In: Proceedings of the IEEE/CVF International Conference on Computer
  Vision (ICCV). pp. 16698--16708 (October 2023)

\bibitem{li2022bigdatasetgan}
Li, D., Ling, H., Kim, S.W., Kreis, K., Fidler, S., Torralba, A.:
  Bigdatasetgan: Synthesizing imagenet with pixel-wise annotations. In:
  Proceedings of the IEEE/CVF Conference on Computer Vision and Pattern
  Recognition. pp. 21330--21340 (2022)

\bibitem{li2022mage}
Li, T., Chang, H., Mishra, S.K., Zhang, H., Katabi, D., Krishnan, D.: Mage:
  Masked generative encoder to unify representation learning and image
  synthesis (2022)

\bibitem{lin2014microsoft}
Lin, T.Y., Maire, M., Belongie, S., Hays, J., Perona, P., Ramanan, D.,
  Doll{\'a}r, P., Zitnick, C.L.: Microsoft coco: Common objects in context. In:
  Computer Vision--ECCV 2014: 13th European Conference, Zurich, Switzerland,
  September 6-12, 2014, Proceedings, Part V 13. pp. 740--755. Springer (2014)

\bibitem{maji13fine-grained}
Maji, S., Kannala, J., Rahtu, E., Blaschko, M., Vedaldi, A.: Fine-grained
  visual classification of aircraft. Tech. rep. (2013)

\bibitem{misra2020self}
Misra, I., Maaten, L.v.d.: Self-supervised learning of pretext-invariant
  representations. In: Proceedings of the IEEE/CVF Conference on Computer
  Vision and Pattern Recognition. pp. 6707--6717 (2020)

\bibitem{tiny-imagenet}
mnmoustafa, M.A.: Tiny imagenet (2017),
  \url{https://kaggle.com/competitions/tiny-imagenet}

\bibitem{nichol2021improved}
Nichol, A.Q., Dhariwal, P.: Improved denoising diffusion probabilistic models.
  In: International Conference on Machine Learning. pp. 8162--8171. PMLR (2021)

\bibitem{nie2020semi}
Nie, W., Karras, T., Garg, A., Debnath, S., Patney, A., Patel, A.B.,
  Anandkumar, A.: Semi-supervised stylegan for disentanglement learning. In:
  Proceedings of the 37th International Conference on Machine Learning. pp.
  7360--7369 (2020)

\bibitem{Nilsback08}
Nilsback, M.E., Zisserman, A.: Automated flower classification over a large
  number of classes. In: Indian Conference on Computer Vision, Graphics and
  Image Processing (Dec 2008)

\bibitem{noroozi2016unsupervised}
Noroozi, M., Favaro, P.: Unsupervised learning of visual representations by
  solving jigsaw puzzles. In: European conference on computer vision. pp.
  69--84. Springer (2016)

\bibitem{oquab2023dinov2}
Oquab, M., Darcet, T., Moutakanni, T., Vo, H., Szafraniec, M., Khalidov, V.,
  Fernandez, P., Haziza, D., Massa, F., El-Nouby, A., Assran, M., Ballas, N.,
  Galuba, W., Howes, R., Huang, P.Y., Li, S.W., Misra, I., Rabbat, M., Sharma,
  V., Synnaeve, G., Xu, H., Jegou, H., Mairal, J., Labatut, P., Joulin, A.,
  Bojanowski, P.: Dinov2: Learning robust visual features without supervision
  (2023)

\bibitem{pang2022unsupervised}
Pang, B., Zhang, Y., Li, Y., Cai, J., Lu, C.: Unsupervised visual
  representation learning by synchronous momentum grouping (2022)

\bibitem{pathak2016context}
Pathak, D., Krahenbuhl, P., Donahue, J., Darrell, T., Efros, A.A.: Context
  encoders: Feature learning by inpainting. In: Proceedings of the IEEE
  conference on computer vision and pattern recognition. pp. 2536--2544 (2016)

\bibitem{pidhorskyi2020adversarial}
Pidhorskyi, S., Adjeroh, D.A., Doretto, G.: Adversarial latent autoencoders.
  In: Proceedings of the IEEE/CVF Conference on Computer Vision and Pattern
  Recognition. pp. 14104--14113 (2020)

\bibitem{pnvr2023ld}
Pnvr, K., Singh, B., Ghosh, P., Siddiquie, B., Jacobs, D.: Ld-znet: A latent
  diffusion approach for text-based image segmentation. In: Proceedings of the
  IEEE/CVF International Conference on Computer Vision. pp. 4157--4168 (2023)

\bibitem{ramesh2022hierarchical}
Ramesh, A., Dhariwal, P., Nichol, A., Chu, C., Chen, M.: Hierarchical
  text-conditional image generation with clip latents. arXiv preprint
  arXiv:2204.06125  (2022)

\bibitem{razavi2019generating}
Razavi, A., Van~den Oord, A., Vinyals, O.: Generating diverse high-fidelity
  images with vq-vae-2. Advances in neural information processing systems
  \textbf{32} (2019)

\bibitem{rombach2022high}
Rombach, R., Blattmann, A., Lorenz, D., Esser, P., Ommer, B.: High-resolution
  image synthesis with latent diffusion models. In: Proceedings of the IEEE/CVF
  Conference on Computer Vision and Pattern Recognition. pp. 10684--10695
  (2022)

\bibitem{saharia2022photorealistic}
Saharia, C., Chan, W., Saxena, S., Li, L., Whang, J., Denton, E.L.,
  Ghasemipour, K., Gontijo~Lopes, R., Karagol~Ayan, B., Salimans, T., et~al.:
  Photorealistic text-to-image diffusion models with deep language
  understanding. Advances in Neural Information Processing Systems
  \textbf{35},  36479--36494 (2022)

\bibitem{sariyildiz2023fake}
Sariyildiz, M.B., Alahari, K., Larlus, D., Kalantidis, Y.: Fake it till you
  make it: Learning transferable representations from synthetic imagenet
  clones. In: CVPR 2023-IEEE/CVF Conference on Computer Vision and Pattern
  Recognition. pp. 1--11 (2023)

\bibitem{Sauer2021ARXIV}
Sauer, A., Schwarz, K., Geiger, A.: Stylegan-xl: Scaling stylegan to large
  diverse datasets. vol. abs/2201.00273 (2022),
  \url{https://arxiv.org/abs/2201.00273}

\bibitem{shrivastava2016contextual}
Shrivastava, A., Gupta, A.: Contextual priming and feedback for faster r-cnn.
  In: Computer Vision--ECCV 2016: 14th European Conference, Amsterdam, The
  Netherlands, October 11--14, 2016, Proceedings, Part I 14. pp. 330--348.
  Springer (2016)

\bibitem{tang2023dift}
Tang, L., Jia, M., Wang, Q., Phoo, C.P., Hariharan, B.: Emergent correspondence
  from image diffusion. arXiv preprint arXiv:2306.03881  (2023)

\bibitem{tomasev2022pushing}
Tomasev, N., Bica, I., McWilliams, B., Buesing, L., Pascanu, R., Blundell, C.,
  Mitrovic, J.: Pushing the limits of self-supervised resnets: Can we
  outperform supervised learning without labels on imagenet? (2022)

\bibitem{vangansbeke2020scan}
Van~Gansbeke, W., Vandenhende, S., Georgoulis, S., Proesmans, M., Van~Gool, L.:
  Scan: Learning to classify images without labels. In: Proceedings of the
  European Conference on Computer Vision (2020)

\bibitem{Horn_2015_CVPR}
Van~Horn, G., Branson, S., Farrell, R., Haber, S., Barry, J., Ipeirotis, P.,
  Perona, P., Belongie, S.: Building a bird recognition app and large scale
  dataset with citizen scientists: The fine print in fine-grained dataset
  collection. In: Proceedings of the IEEE Conference on Computer Vision and
  Pattern Recognition (CVPR) (June 2015)

\bibitem{WahCUB_200_2011}
Wah, C., Branson, S., Welinder, P., Perona, P., Belongie, S.: {The Caltech-UCSD
  Birds-200-2011 Dataset}. Tech. Rep. CNS-TR-2011-001, California Institute of
  Technology (2011)

\bibitem{walmer2023teaching}
Walmer, M., Suri, S., Gupta, K., Shrivastava, A.: Teaching matters:
  Investigating the role of supervision in vision transformers (2023)

\bibitem{Xiang_2023_ICCV}
Xiang, W., Yang, H., Huang, D., Wang, Y.: Denoising diffusion autoencoders are
  unified self-supervised learners. In: Proceedings of the IEEE/CVF
  International Conference on Computer Vision (ICCV). pp. 15802--15812 (October
  2023)

\bibitem{xiao2018unified}
Xiao, T., Liu, Y., Zhou, B., Jiang, Y., Sun, J.: Unified perceptual parsing for
  scene understanding. In: Proceedings of the European conference on computer
  vision (ECCV). pp. 418--434 (2018)

\bibitem{xu2023openvocabulary}
Xu, J., Liu, S., Vahdat, A., Byeon, W., Wang, X., Mello, S.D.: Open-vocabulary
  panoptic segmentation with text-to-image diffusion models (2023)

\bibitem{yin2019automatic}
Yin, C., Qian, B., Wei, J., Li, X., Zhang, X., Li, Y., Zheng, Q.: Automatic
  generation of medical imaging diagnostic report with hierarchical recurrent
  neural network. In: 2019 IEEE international conference on data mining (ICDM).
  pp. 728--737. IEEE (2019)

\bibitem{yu2021vector}
Yu, J., Li, X., Koh, J.Y., Zhang, H., Pang, R., Qin, J., Ku, A., Xu, Y.,
  Baldridge, J., Wu, Y.: Vector-quantized image modeling with improved vqgan.
  In: International Conference on Learning Representations (2021)

\bibitem{pmlr-v139-zbontar21a}
Zbontar, J., Jing, L., Misra, I., LeCun, Y., Deny, S.: Barlow twins:
  Self-supervised learning via redundancy reduction. In: Meila, M., Zhang, T.
  (eds.) Proceedings of the 38th International Conference on Machine Learning.
  Proceedings of Machine Learning Research, vol.~139, pp. 12310--12320. PMLR
  (18--24 Jul 2021), \url{https://proceedings.mlr.press/v139/zbontar21a.html}

\bibitem{zhang2023tale}
Zhang, J., Herrmann, C., Hur, J., Cabrera, L.P., Jampani, V., Sun, D., Yang,
  M.H.: A tale of two features: Stable diffusion complements dino for zero-shot
  semantic correspondence  (2023)

\bibitem{zhang2016colorful}
Zhang, R., Isola, P., Efros, A.A.: Colorful image colorization. In: European
  conference on computer vision. pp. 649--666. Springer (2016)

\bibitem{zhang2021datasetgan}
Zhang, Y., Ling, H., Gao, J., Yin, K., Lafleche, J.F., Barriuso, A., Torralba,
  A., Fidler, S.: Datasetgan: Efficient labeled data factory with minimal human
  effort. In: Proceedings of the IEEE/CVF Conference on Computer Vision and
  Pattern Recognition. pp. 10145--10155 (2021)

\bibitem{zhao2023unleashing}
Zhao, W., Rao, Y., Liu, Z., Liu, B., Zhou, J., Lu, J.: Unleashing text-to-image
  diffusion models for visual perception (2023)

\bibitem{zhou2017scene}
Zhou, B., Zhao, H., Puig, X., Fidler, S., Barriuso, A., Torralba, A.: Scene
  parsing through ade20k dataset. In: Proceedings of the IEEE conference on
  computer vision and pattern recognition. pp. 633--641 (2017)

\bibitem{zhou2022ibot}
Zhou, J., Wei, C., Wang, H., Shen, W., Xie, C., Yuille, A., Kong, T.: ibot:
  Image bert pre-training with online tokenizer (2022)

\bibitem{zhou2022mugs}
Zhou, P., Zhou, Y., Si, C., Yu, W., Ng, T.K., Yan, S.: Mugs: A multi-granular
  self-supervised learning framework (2022)

\end{thebibliography}

% Supplementary

\clearpage
\maketitlesupplementary

\section{Diffusion Models Fundamentals}
\label{sec:diffusion-fundamentals-full}
Diffusion models first define a forward noising process where Gaussian noise is iteratively added to an image $x_0$ sampled from the data distribution $q(x_0)$, to get a completely noised image $x_T$ in $T$ steps. 
This forward process is defined as a Markov chain with noisy image latents $x_1,x_2 \dots, x_t, \dots,x_{T-1},x_T$.
Formally, the forward diffusion process is
\begin{equation}
\label{eq:forward}
\begin{split}
q(x_1,\dots x_T|x_0) &:= \prod_{t=1}^{T}q(x_t|x_{t-1}) \\
q(x_t|x_{t-1}) &:= \mathcal{N} (x_t; \sqrt{1-\beta_t}x_{t-1}, \beta_t\textbf{I})
\end{split}
\end{equation}
where ${\{\beta_t\}}_{t=1}^T$ is the variance schedule and $\mathcal{N}$ is a normal distribution. 
As $T\rightarrow\infty$, $x_T$ nearly is equivalent to the isotropic Gaussian distribution. With $\alpha_t:=1-\beta_t$ and $\Bar{\alpha}_t:=\prod_{i=0}^t\alpha_i$ one can sample a noised image $x_t$ at diffusion step $t$ directly from a real image $x_0$ using Eq.~\ref{eq:make_noisy}.

The reverse diffusion process aims to reverse the forward process and sample from the posterior distribution $q(x_{t-1}|x_{t})$ which depends on the entire data distribution.  
Doing this iteratively can denoise a completely noisy image $x_T$, such that one can sample from the data distribution $q(x_0)$. 
This is approximated using a neural network $\epsilon_\theta$ as   
\begin{equation}
\label{eq:reverse}
\begin{split}
p_{\theta}(x_{t-1}|x_{t}) :=
\mathcal{N} \left( x_{t-1}; \frac{x_t-\frac{\beta_t}{\sqrt{1-\Bar{\alpha}_t}}\epsilon_\theta(x_t,t)}{\sqrt{\alpha_t}},\Sigma_\theta(x_t,t) \right) 
\end{split}
\end{equation}

When $p$ and $q$ are interpreted as a VAE, a simplified version of the variational lower bound objective turns out to be just a mean squared error loss~\cite{ho2020denoising}. This can be used to train $\epsilon_\theta$ which learns to approximate the Gaussian noise $\epsilon$ added to the real image $x_0$ in Eq. \ref{eq:make_noisy} as
\begin{equation}
\label{eq:loss}
\begin{split}
\mathcal{L}_{\text{simple}}=\mathbb{E}_{x_0,t,\epsilon}[\|\epsilon_\theta(x_t, t) - \epsilon \|_2^2]
\end{split}
\end{equation}
$\Sigma_\theta(x_t,t)$ is either kept fixed~\cite{ho2020denoising} or is learned using the original variational lower-bound objective~\cite{nichol2021improved,dhariwal2021diffusion}. 

\section{Training Details}
\label{sec:training_details}
In this section, we provide training details for our benchmarking over various tasks. We provide our code in the supplementary bundle. Please refer to the code for exact details of our methods and their training. 
\subsection{Datasets and Parameters}

\begin{table}[h]
    \begin{minipage}{.58\linewidth}
\caption{Classification dataset details.}
\label{tab:datasets}
\centering
    \resizebox{\textwidth}{!}{
    \setlength{\tabcolsep}{10pt}
    \begin{tabular}{@{}l c c c@{}} 
        \toprule
        Dataset & \multicolumn{1}{c}{\#Cls} & \multicolumn{1}{c}{\#Train} & \multicolumn{1}{c}{\#Test} \\
        \midrule
        Aircraft \cite{maji13fine-grained} & 100 & 6,667 & 3,333 \\
        Cars \cite{KrauseStarkDengFei-Fei_3DRR2013} & 196 & 8,144 & 8,041 \\
        CUB \cite{WahCUB_200_2011} & 200 & 5,994 & 5,794 \\
        Dogs \cite{KhoslaYaoJayadevaprakashFeiFei_FGVC2011} & 120 & 12,000 & 8,580 \\
        Flowers \cite{Nilsback08} & 102 & 2,040 & 6,149 \\
        NABirds \cite{Horn_2015_CVPR} & 555 & 23,929 & 24,633 \\
        ImageNet-1k \cite{deng2009imagenet} & 1000 & 1.28M & 50,000 \\
        ImageNet 1\% labels \cite{chen2020simple} & 1000 & 12811 & 50,000 \\
        ImageNet 10\% labels \cite{chen2020simple} & 1000 & 128,116 & 50,000 \\
        ImageNet-50 \cite{vangansbeke2020scan} & 50 & 64,274 & 2,500 \\
        \bottomrule
    \end{tabular}
}
    \end{minipage}
    \hfill
    \begin{minipage}{.36\linewidth}
\begin{center}
\caption{Parameter counts of major unified unsupervised representation learning methods. For each, we consider the whole system, not just the encoding network.}
\label{tab:parameter-counts}
    \resizebox{.7\textwidth}{!}{
    \setlength{\tabcolsep}{8pt}
\begin{tabular}{@{}l|c@{}}
\toprule
Method & \# Params  \\
\midrule
BigBiGAN & 502M  \\
MAGE & 439M \\
GD & 553M \\
DifFormer &585M \\
DifFeed &583M \\
\bottomrule
\end{tabular}
}
\end{center}
    \end{minipage} 
\end{table}

For classification, we use the datasets shown in Table~\ref{tab:datasets}. 
For object detection and semantic segmentation, COCO~\cite{lin2014microsoft} dataset was used which consists of 118K training and 5K validation images containing 1.5M object instances over 80 object categories. 
For semantic segmentation, we use the SceneParse150 benchmark of ADE20K~\cite{zhou2017scene} dataset which contains 20,210 training and 3,169 validation images having 150 categories which occupy the most pixels in the images.

Please find the details of the parameter counts of various  unsupervised unified representation learners in Table~\ref{tab:parameter-counts}.

\subsection{Semi-supervised Details}

As with our linear probing experiments, we follow VISSL protocols for the semi-supervised experiments -- refer to their code for exact hyperparameter settings.
Key settings, such as data augmentations, classification head, and features selected, are the same as for linear probing.
However, since we full-finetune the model (both ours and baselines) for this setting, the optimizer, learning rate, and number of epochs are changed to SGD, 0.05, and 70, respectively, with a single step reduction of the learning rate to 0.005 at 30 epochs.
The batch size remains 256, and wherever this has changed, the learning rate should be inversely scaled accordingly.

\subsection{Object Detection Details}

For all object detection results in Figure~\ref{fig:detection}, we train on the equal number of epochs -- 10.
We scale the step learning rate schedulers for ViT and ResNet accordingly, and borrow the ViT settings for the guided diffusion backbone.
We use a batch size of 4 for both ViT and diffusion (1 per GPU), and choose learning rate based on a hyperparameter search for both models.
With ResNet, we are able to use the original batch size (16).

\subsection{Semantic Segmentation Details}
The official semantic segmentation code for MAE has not been made available. Hence, we use mae\_segmentation repository (\url{https://github.com/implus/mae_segmentation}) for the settings for MAE (ViT-B) on MMSegmentation framework which reproduces the results from the MAE paper. It uses 2 conv layers, 1 conv-transpose layer, identity, a maxpool layer respectively to convert their embeddings from different layers to UperNet inputs.  
We use batch size 2 due to resource constraints (instead of 16) in all the models we train. We tried tuning the learning rate for updated batch size and the default worked the best. For MAE (ViT-L), we update the hyperparameters according to ViT-Det COCO Object Detection following a comment by the MAE authors \href{https://github.com/facebookresearch/mae/issues/52#issuecomment-1095406286}{comment by the MAE authors}
and increase the UperNet channel dimension to 1024 instead of 768 for MAE (ViT-B). 
For GD, we convert the GD U-Net's features to 1024 using 1$\times$1 conv layer and then use bilinear interpolation to get the shapes of features identical to that of MAE (ViT-L)'s inputs to the UperNet. 
We also performed learning rate and block-choice hyperparameter tuning. We finally use the learning rate 2e-5. 
Please note that the ResNet50 backbone results are picked from DreamTeacher~\cite{Li_2023_ICCV} which might have different hyperparameter settings given their code is unavailable (e.g. they use the default batch size 16).

\section{Additional Results}
\subsection{K-Nearest Neighbor Results}

\begin{table*}[h]
\begin{minipage}[t]{0.61\linewidth}
\begin{center}
\caption{kNN Results on ImageNet-1k.}
\label{tab:knn-results}
    \resizebox{\textwidth}{!}{
    \setlength{\tabcolsep}{10pt}
    \begin{tabular}{@{}c c c | c c c c@{}} 
        \toprule
        b & t & pool & Top1@20 & Top5@20 & Top1@200 & Top5@200 \\
        \midrule
        19 & 10  & 1 & 41.14 & 62.49 & 39.04 & 65.40 \\
        19 & 30  & 1 & 46.30 & 68.47 & 43.65 & 70.94 \\
        19 & 50  & 1 & 47.89 & 70.10 & 45.06 & 72.36 \\
        19 & 90  & 1 & 49.04 & 71.40 & 46.17 & 73.65 \\
        19 & 150 & 1 & 48.78 & 71.27 & 46.29 & 73.76 \\
        19 & 300 & 1 & 45.18 & 67.28 & 43.04 & 70.09 \\
        19 & 500 & 1 & 29.96 & 48.64 & 29.68 & 53.12 \\
        24 & 90  & 1 & 48.34 & 70.02 & 46.22 & 72.57 \\
        \bottomrule
    \end{tabular}
}
\end{center}
\end{minipage}
\hfill
\begin{minipage}[t]{0.35\linewidth}
\caption{Stable Diffusion linear probe results.}%
\label{tab:stable-diffusion}
\centering
    \resizebox{\textwidth}{!}{
    \setlength{\tabcolsep}{10pt}
    \begin{tabular}{@{}l c c | c c@{}} 
        \toprule
        Condition & $b$ & Size & Accuracy \\
        \midrule
        Null Text  & 18 & 512  & 64.67\% \\
        Null Text  & 15 & 512 & 55.77\% \\
        Null Text  & 18 & 256  & 41.37\%\\
        Learnable  & 18 & 512    & \textbf{65.18\%} \\
        \midrule
        GD & 24 & 256 & \textbf{61.86 \%} \\
        \bottomrule
    \end{tabular}
}
\end{minipage}
\end{table*}

Apart from linear probing results in the main paper we provide kNN results in Table~\ref{tab:knn-results}.  We observe the trends exactly match that in Figure~\ref{fig:bn_t_acc} from the main paper. There are minor deviations, the most significant being that $t=150$, $b=24$ is now the best setting by a slightly larger margin (1\%), but the trends support the same conclusion we draw w.r.t. feature selection.

\subsection{Stable Diffusion features}

As a proof of concept, we also test our hypothesis on Stable Diffusion (SD) features trained on ImageNet-1k for 15 epochs. Note that Stable Diffusion has been trained on paired text-image data, and hence has indirectly seen a lot of labels in the form of text captions, and hence can not be called unsupervised. Nonetheless, we show the results in Table \ref{tab:stable-diffusion}) as an alternative diffusion model to see the generalizability of diffusion-based pertaining. For text conditioning we try both null text as well as learnable text embedding. We see that SD U-Net's mid-block $b=15$ under-performs $b=18$ which is consistent with GD. Further, we see that a larger image size and learnable condition are better. Our results are also consistent with SD Features baseline in \cite{li2023your} at a comparable setting, but this shows that SD Features' performance can be further improved by optimizing the choice of block (to an accuracy better than their zero-shot performance).

\subsection{More CKA comparisons}
\begin{figure}[!h]
    \centering
\includegraphics[width=0.8\linewidth]{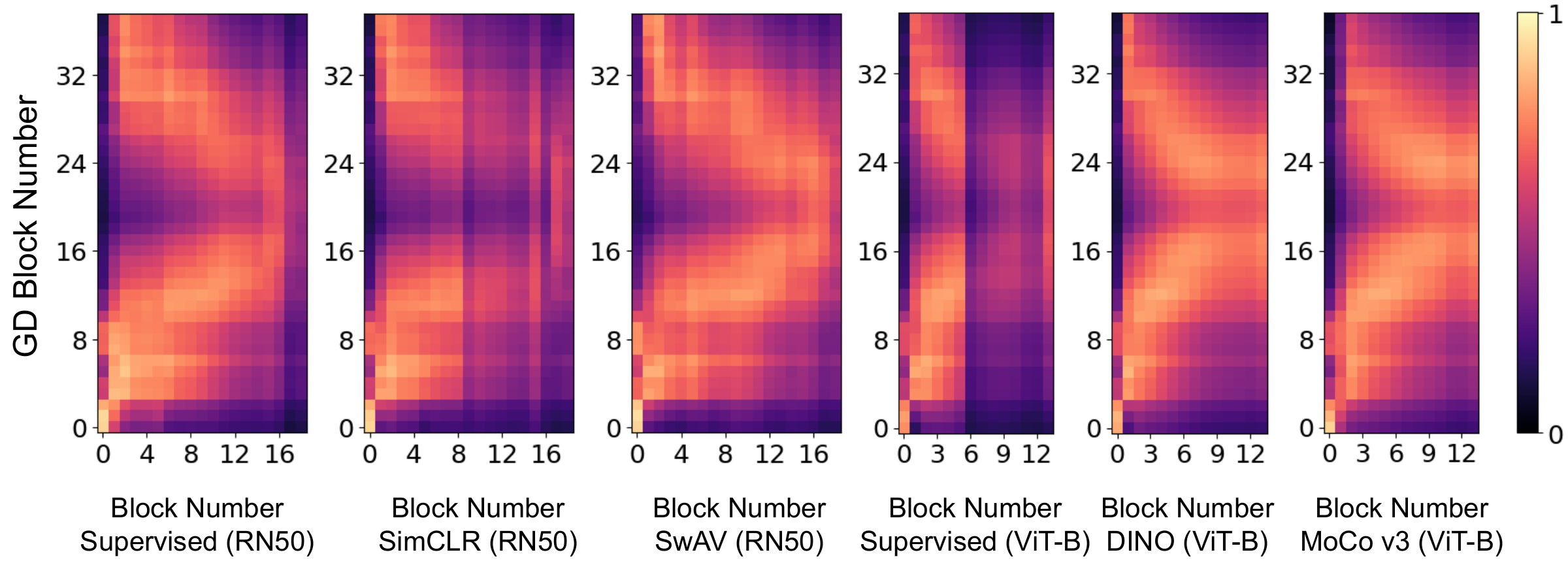}
    \caption{Feature representation comparisons via centered kernel alignment (CKA). RN50 stands for ResNet-50. }
    \label{fig:cka_more_3}
\end{figure}

One of our key findings is that unconditional diffusion training can be considered as a self-supervised pretraining strategy. Hence, similar to~\cite{Gwilliam_2022_CVPR,walmer2023teaching}, we compare their representations in Figure~\ref{fig:cka_analysis}(a) (MAE) and offer additional comparisons (MoCo-v3, etc.) in Figure~\ref{fig:cka_more_3}.

\section{Ablations}

\subsection{Attention Head Alternatives}

\begin{table*}[h]
\begin{minipage}[t]{0.43\linewidth}
\captionof{figure}{Head results. We show the results of explorations for various classification heads -- Linear, MLP, CNN, Attention, trained on frozen features for ImageNet-50, computed at block number 24 and noise time step 90.}
    \includegraphics[width=1.\linewidth, trim={14.2cm 0 0 0.69cm},clip]{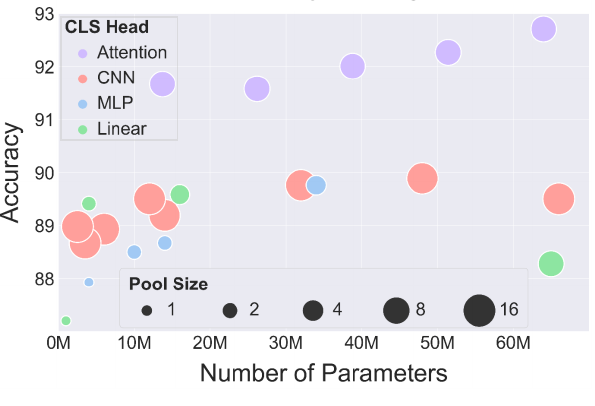}
    
    \label{fig:head_comparison}
\end{minipage}
\hfill
\begin{minipage}[t]{0.53\linewidth}
\caption{Linear and MLP results. For linear, 1k, 4k, 16k, and 65k indicate the size of the feature after pooling and flattening. For MLP, the first number is the size of the feature after pooling and flattening, and the succeeding numbers are hidden sizes of layers before the last (classification) layer.}
\label{tab:lin-mlp-results}
\centering
    \resizebox{.75\textwidth}{!}{
    \setlength{\tabcolsep}{10pt}
    \begin{tabular}{@{}l | c c@{}} 
        \toprule
        Head & Params & Accuracy \\
        \midrule
        Linear-1k    & 1M & 87.20\% \\
        Linear-4k    & 4M & 89.41\% \\
        Linear-16k   & 16M & 89.58\% \\
        Linear-65k   & 65M & 88.28\% \\
        MLP-1k-2k    & 4M & 87.93\% \\
        MLP-4k-2k    & 10M & 88.50\% \\
        MLP-4k-2k-2k & 14M & 88.67\% \\
        MLP-16k-2k   & 34M & 89.76\% \\
        \bottomrule
    \end{tabular}
}
\end{minipage}
\end{table*}

\begin{table*}[htb]
\begin{minipage}[t]{0.56\linewidth}
\caption{CNN head results. Channel sizes are separated by dashes. The first is the channel size of the input feature maps, the next is the output channels from the first convolution, and the last is the output dimension of the second convolution. These are treated as hyperparameters. For more detail, see our code.}
\label{tab:cnn-results}
\centering
    \resizebox{.8\textwidth}{!}{
    \setlength{\tabcolsep}{8pt}
    \begin{tabular}{@{}l|ccc@{}}
        \toprule
        Head & Params & Accuracy \\
        \midrule
        CNN-1k-256-256    & 2.5M & 88.98\% \\
        CNN-1k-512-256    & 3.5M & 88.67\% \\
        CNN-1k-1k-256   & 6M & 88.93\% \\
        CNN-1k-1k-1k   & 12M & 89.50\% \\
        CNN-1k-2k-512    & 14M & 89.19\% \\
        CNN-1k-2k-2k     & 32M & 89.76\% \\
        CNN-1k-4k-2k  & 48M & 89.89\% \\
        CNN-1k-4k-2.5k   & 66M & 89.50\% \\
        \bottomrule
\end{tabular}
}
\end{minipage}
\hfill
\begin{minipage}[t]{0.41\linewidth}
\caption{Attention head results. The hyperparameters are denoted following the dashes. The first hyperparameter is similarly the channel size of the input feature maps and the next is the number of Transformer blocks used. For fair comparison with other heads, we remove $1 \times 1$ convolution in tokenizer for this analysis.}
\label{tab:attention-results}
\centering
    \resizebox{0.95\textwidth}{!}{
    \setlength{\tabcolsep}{10pt}
    \begin{tabular}{@{}l | c c@{}} 
        \toprule
        Head & Params & Accuracy \\
        \midrule
        Attention-1K-1 & 13.7M & 91.67\% \\
        Attention-1K-2 & 26.2M & 91.58\% \\
        Attention-1K-3 & 38.8M & 92.01\% \\
        Attention-1K-4 & 51.4M & 92.27\% \\
        Attention-1K-5 & 64.0M &  92.71\% \\
        \bottomrule
    \end{tabular}
}
\end{minipage}
\end{table*}

Apart from the Attention head presented in the main paper, we also experimented with other heads like Linear, MLP, and CNN of various powers on ImageNet-50 as seen in Figure~\ref{fig:head_comparison}. More information is present in Table~\ref{tab:lin-mlp-results}, Table~\ref{tab:cnn-results}, and Table~\ref{tab:attention-results}.
We show exact accuracies and parameter counts.
We also specify our hyperparameter selection for each head. It can be clearly seen that at the same level of parameters Attention head outperforms the other alternatives.

\subsection{Feedback Feature Extraction}

\begin{table}[h!]
\centering
\caption{
DifFeed feature extraction ablation. 
}
\label{tab:feedback-secondpass}
\resizebox{0.8\linewidth}{!}{
\setlength{\tabcolsep}{10pt}
\begin{tabular}{@{}l| c| c c c c c c@{}}
\toprule
 \multicolumn{2}{r}{$b$}& 21 & 24 & 27& 30 & 33& 36 \\
\midrule
\multirow{2}{*}{Accuracy} & CUB & 65.1\% & \textbf{71.0\%} & 70.6\% & 67.2\% & 56.6\% & 49.8\%\\
& Cars & 68.3\% & \textbf{82.7\%}& 73.3\% & - & - & -\\
\bottomrule

\end{tabular}
}

\end{table}

We provide the ablation for various strategies for feedback mechanisms in DifFeed in Section ~\ref{subsec:ablations} of the main paper. Here we provide our ablation for the choice of $b=24$ for extracting the final feature in the second pass in Table~\ref{tab:feedback-secondpass}. The results are presented for the ``bottleneck'' strategy of feedback.

\subsection{Transformer Feature Fusion comparisons}

\begin{table}[h]
\centering
\begin{minipage}{0.8\linewidth}
\begin{center}
\caption{Transformer feature fusion results.}
\label{tab:fusion-head-comparison}
    \resizebox{\textwidth}{!}{
    \setlength{\tabcolsep}{8pt}
\begin{tabular}{@{}l c|c c | c c@{}}
\toprule
Method & Backbone &$\mathcal{T}$ & $\mathcal{B}$ & Params & Accuracy  \\
\midrule
SimCLR & ResNet-50 & - & \{4\}$^\dagger$ & 27.5M & 93.6\%\\
SimCLR & ResNet-50 & - & \{1,2,3,4\}* & 30.0M & 93.6\%\\
\midrule
MAE & ViT-B & - & \{12\}$^\dagger$ & 26.0M & 93.6\%\\
MAE & ViT-B & - & \{9,10,11,12\}* & 28.4M & 94.1\%\\
\midrule
Attention & GD & \{150\} & \{24\} &  26.8M & 91.1\%\\
DifFormer & GD & \{90,150,300\} & \{19,24,30\} &  29.3M & 94.2\%\\
DifFeed & GD & \{150\} & \{24\} &  29.6M &\textbf{94.9}\%\\
\bottomrule

\end{tabular}
}
\hspace{-.02in}
\footnotesize{$^\dagger$Attention head is applied.\\ *Transformer feature fusion mechanism is applied.} \\

\end{center}
\end{minipage}
\end{table}

For fair comparison in terms of the number of learnable parameters, we provide results of applying our Attention head and transformer feature fusion mechanisms on top of baselines . We number blocks in ResNet-50 as $b \in \{1, 2, 3, 4\}$ based on the feature map size and blocks in ViT-B as $b \in \{1, 2, ..., 12\}$ based on the attention layers. We fixed the backbones and trained heads on ImageNet-50 for 28 epochs and compare them against our approaches. The results for the best learning rate for each of the methods are provided in Table~\ref{tab:fusion-head-comparison}. We see that transformer feature fusion mechanism does not help SimCLR~\cite{chen2020simple} but has slight improvement in MAE~\cite{DBLP:journals/corr/abs-2111-06377}. In the case of our methods, we see that feature fusion and feedback mechanisms provide performance boosts of more than 3\% over just using Attention head and consequently outperform both the baselines with or without feature fusion. This shows the efficacy of both these mechanisms in the case of guided diffusion.

\subsection{Inference Times and Additional times/blocks}

\begin{table}[h]
    \begin{minipage}{.48\linewidth}
      \caption{\footnotesize Avg.\ inference time (sec), w/ batch size 1. We also provide the number of forward passes through the backbone. }
\label{tab:runtime}
      \centering
      \resizebox{\textwidth}{!}{
      \setlength\tabcolsep{4pt}
\begin{tabular}{@{}l|cccc@{}} 
    \toprule
    Method & GD & Attention & DifFormer & DifFeed \\
    \midrule
    time (sec) & 0.072 & 0.075 & 0.228 & 0.146 \\
    \# Forwards & 1 & 1 & 3 & 2\\
    \bottomrule
\end{tabular}
    }
    \end{minipage}%
    \hfill
    \begin{minipage}{.02\linewidth}
    \end{minipage}
    \begin{minipage}{.48\linewidth}
    \caption{\footnotesize CUB performance (vs.\ default) with additional times and blocks.}
\label{tab:additional_t_b}
      \centering

    \resizebox{0.9\textwidth}{!}{
    \footnotesize
    \setlength\tabcolsep{4pt}
        \begin{tabular}{@{}lcc@{}} 
        \toprule
        Method & Accuracy  & \#Fwds \\
        \midrule
        DifFormer (7 $t$s, 7 $b$s)
        
        & 72.0\% (--0.4\%) & 7 (+4)   \\
        \midrule
        DifFeed (3 $t$s)
        & 74.4\% (+3.4\%) & 6 (+4)   \\
        \bottomrule
    \end{tabular}
    }
    \end{minipage} 
\end{table}

We provide inference runtime comparisons in Table~\ref{tab:runtime}. We see that runtimes are directly proportional to the number of forward passes through the GD backbone, which consequently shows DifFeed's efficiency over DifFormer. Furthermore, in Table \ref{tab:additional_t_b} we explore the use of more times/blocks in our DifFormer and DifFeed on the CUB dataset for FGVC. For DifFormer we tried using features from 7 blocks over 7 timesteps ($\mathcal{B}=\{6,9,15,19,24,30,33\}$, $\mathcal{T}=\{10,30,50,90,150,300,500\}$) which require 7 forward passes. For DifFeed we tried to provide feedback from 3 different timesteps ($\mathcal{T}=\{90,150,300\}$) each of which requires 2 forwards, hence a total of 6 forward passes. We see that using additional times and blocks has noticeable, but diminishing returns compared to our default settings.

\end{document}